\documentclass[10pt,twocolumn,letterpaper]{article}

\usepackage{iccv}
\usepackage{times}
\usepackage{epsfig}
\usepackage{graphicx}
\usepackage{amsmath}
\usepackage{amssymb}
\usepackage{bm}

\usepackage{grffile}
\usepackage{multirow}
\usepackage{mathtools}
\DeclarePairedDelimiterX{\inp}[2]{\langle}{\rangle}{#1, #2}

\usepackage{tabularx}

\usepackage[british,UKenglish]{babel}

\newcommand{\figref}[1]{Fig.~\ref{#1}}
\newcommand{\tblref}[1]{Table~\ref{#1}}
\newcommand{\sref}[1]{Sect.~\ref{#1}}

\newcommand{\real}{\mathbb{R}}
\newcommand{\bx}{\mathbf{x}}

\newcommand{\argmax}{\operatornamewithlimits{argmax}}

\newcommand{\bI}[0]{\mathbf{I}}

\newcommand{\mR}[0]{\mathbb{R}}

\usepackage{color}

\usepackage{pax}

 \usepackage{booktabs} 
\usepackage{setspace}
\usepackage{subfigure}
\usepackage{paralist}
\usepackage[numbers,sort]{natbib} 
 
\usepackage[pagebackref=true,breaklinks=true,letterpaper=true,colorlinks,bookmarks=false]{hyperref}

 \iccvfinalcopy 

\def\iccvPaperID{1806} 

\ificcvfinal\fi

\author{Christoph Feichtenhofer\thanks{Christoph \mbox{Feichtenhofer} is a recipient of a DOC Fellowship of the
        Austrian Academy of Sciences at the Institute of Electrical
        Measurement and Measurement Signal Processing, Graz University of
        Technology.}
    \\Graz University of Technology\\ {\tt \small \href{mailto:feichtenhofer@tugraz.at}{\textcolor{black}{feichtenhofer@tugraz.at}}} \and	Axel Pinz\\Graz University of Technology\\{\tt \small \href{mailto:axel.pinz@tugraz.at}{\textcolor{black}{axel.pinz@tugraz.at}}} \and Andrew Zisserman\\
	University of Oxford\\{\tt \small  \href{mailto:az@robots.ox.ac.uk}{\textcolor{black}{az@robots.ox.ac.uk}}}
	}
\begin{document}

\title{Detect to Track and Track to Detect}

\maketitle
\thispagestyle{empty}

\begin{abstract}
	
Recent approaches for high accuracy detection and tracking of object
categories in video consist
of complex multistage solutions that become more cumbersome each
year.  In this paper we propose a ConvNet architecture that jointly
performs detection and tracking, solving the task in a simple and
effective way.

Our contributions are threefold: (i) we set up a ConvNet architecture for simultaneous detection and tracking, using a multi-task objective for frame-based object detection and across-frame track regression; (ii) we introduce correlation features that represent object co-occurrences across time to aid the ConvNet during tracking;  and (iii) we link the frame level detections based on our across-frame tracklets to produce high accuracy detections
at the video level. Our ConvNet architecture for spatiotemporal
object detection is evaluated on the large-scale ImageNet VID dataset
where it achieves state-of-the-art results. Our approach provides
better single model performance than the winning method of the last
ImageNet challenge while being conceptually much simpler. Finally, we
show that by increasing the temporal stride we can dramatically
increase the tracker speed. Our code and models are available at \href{http://github.com/feichtenhofer/detect-track}{\small \texttt{http://github.com/feichtenhofer/detect-track}}

\end{abstract}

\section{Introduction}
\label{sec:intro}
Object detection in images has received a lot of attention over the last
years with tremendous progress mostly due to the emergence of deep
Convolutional Networks
\cite{Lecun89,Krizhevsky12,He16,Szegedy15,simonyan2014very} and their
region based descendants
\cite{Girshick14,Girshick15,ren2016faster,rfcnNIPS16}. 
In the case of object detection and tracking in videos, recent approaches
have mostly used detection as a first step, followed by 
post-processing methods such as applying a tracker to propagate
detection scores over time. Such variations on the `tracking by detection' 
paradigm have seen impressive progress
but are dominated by frame-level detection methods.

\begin{figure}[!t]
	\centering
	\subfigcapskip=-15pt
	\subfigtopskip=-1pt
	\subfigure[]{
		\includegraphics[width=0.22\textwidth]{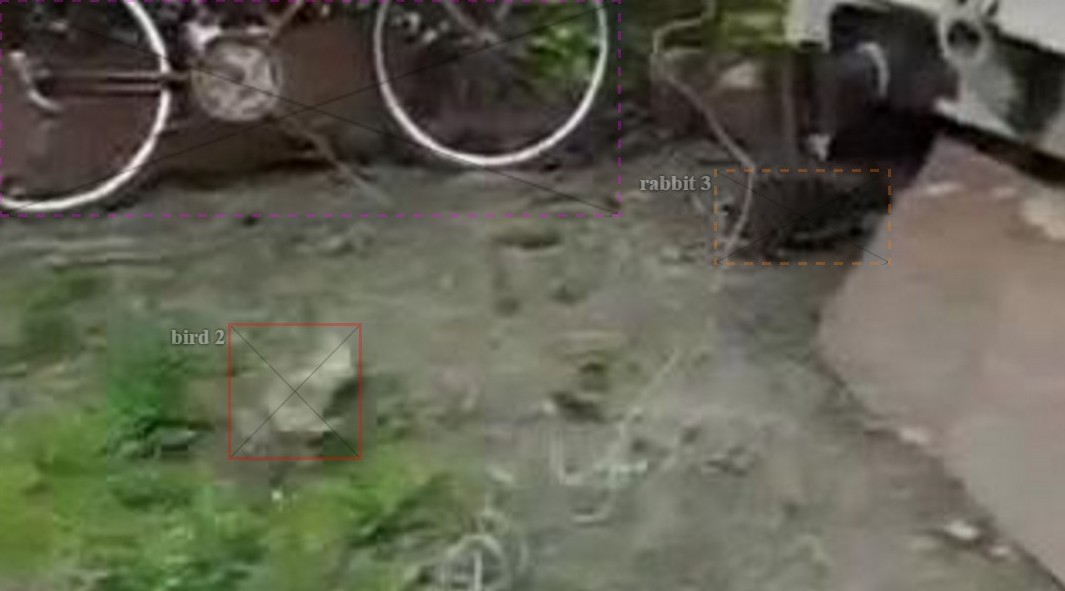}   
		\label{fig:rabbit}}
	\subfigure[]{
		\includegraphics[width=0.22\textwidth]{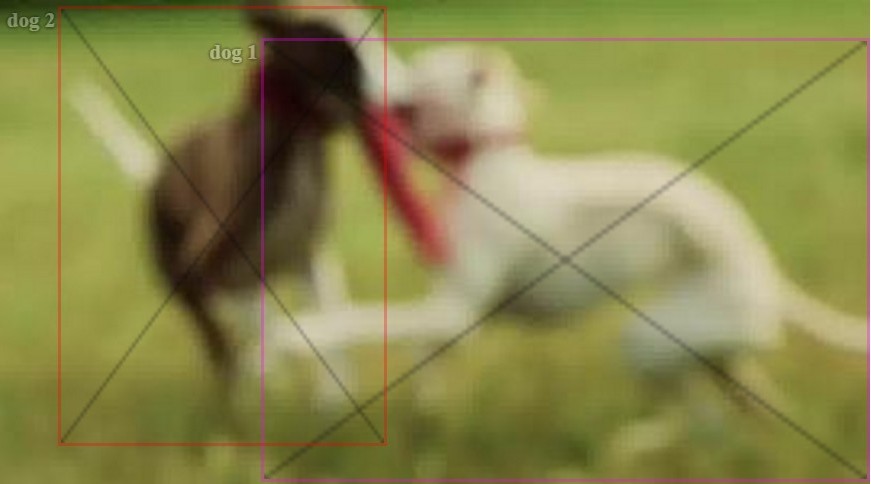}   
		\label{fig:dog}}
	\subfigure[]{
		\includegraphics[width=0.22\textwidth]{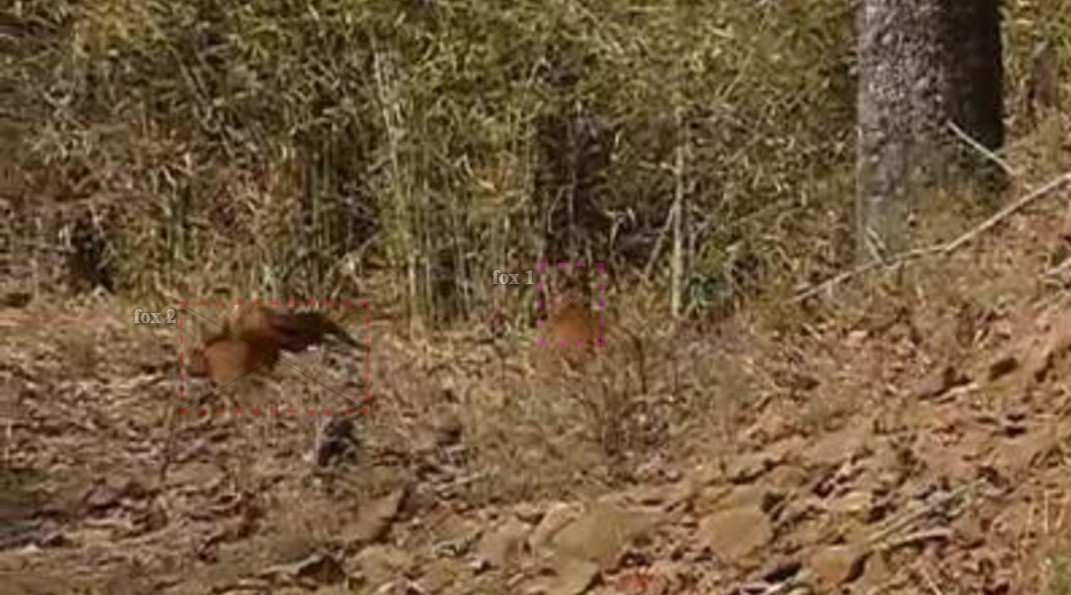}   
		\label{fig:fox}}
	\subfigure[]{
		\includegraphics[width=0.22\textwidth]{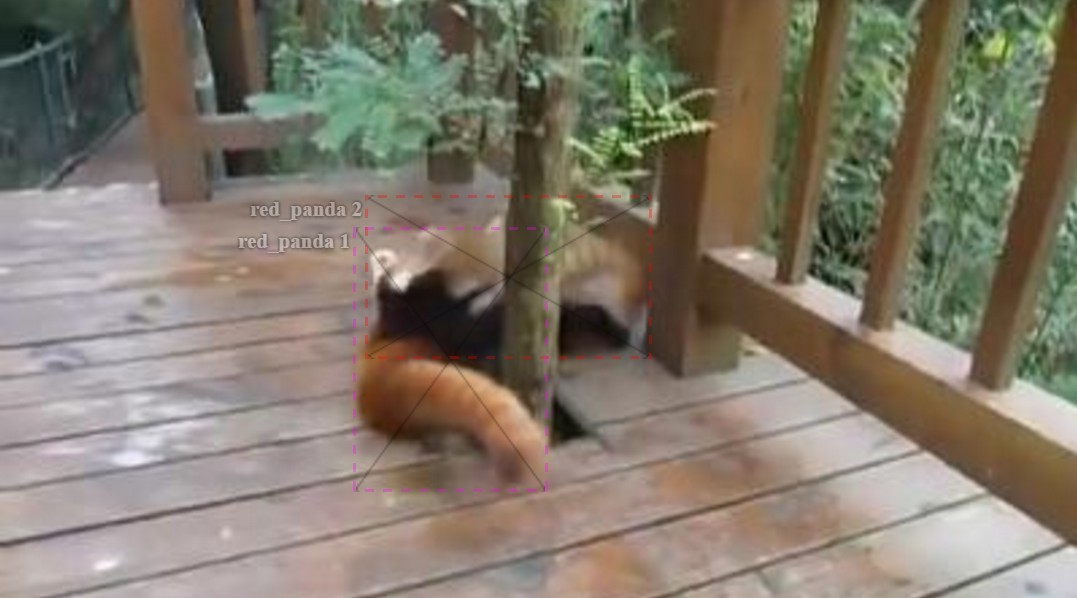}   
		\label{fig:red_panda}}
	\vspace{-2pt}
	\caption{Challenges for video object detection. 
Training examples for: \subref{fig:rabbit}
 bicycle, bird, rabbit; \subref{fig:dog} dog; \subref{fig:fox} fox;  and 
\subref{fig:red_panda} red panda.
	}
\vspace{-10pt}
	\label{fig:defects}
\end{figure}

Object detection in video has seen a surge in interest lately,
especially since the introduction of the ImageNet \cite{ILSVRC15}
video object detection challenge (VID). Different from the ImageNet
object detection (DET) challenge, VID shows objects in image sequences
and comes with additional challenges of (i) size: the sheer number of frames that video provides
(VID has around 1.3M images, compared to around
400K in DET or 100K in COCO \cite{lin2014microsoft}), (ii) motion
blur: due to rapid camera or object motion, (iii) quality: internet
video clips are typically of lower quality than static photos, (iv)
partial occlusion: due to change in objects/viewer positioning, and (v)
pose: unconventional object-to-camera poses are frequently seen in
video. In \figref{fig:defects}, we show example images from the VID
dataset; for more examples please
see\footnote{\url{http://vision.cs.unc.edu/ilsvrc2015/ui/vid}}.

To solve this challenging task, recent top entries in the ImageNet \cite{ILSVRC15} video detection challenge use exhaustive post-processing on top of frame-level detectors. For example, the winner \cite{kang2016tcnn} of ILSVRC'15 uses two multi-stage Faster R-CNN \cite{ren2016faster} detection frameworks, context suppression, multi-scale training/testing, a ConvNet tracker \cite{wang2015visual},  optical-flow based score propagation and model ensembles.

In this paper we propose a unified approach to tackle the problem of object detection in realistic video. 
Our objective is to directly infer a `tracklet' over
multiple frames by simultaneously carrying out detection and tracking
with a ConvNet. To achieve this we propose to extend the R-FCN \cite{rfcnNIPS16} detector with a tracking formulation that is inspired by current correlation and regression based trackers~\cite{bertinetto2016fully,ma2015hierarchical,held2016learning}. We train a fully convolutional architecture end-to-end using a detection and tracking based loss and term our approach D\&T for joint Detection and Tracking. The input to the network consists of multiple frames which are first passed through a ConvNet trunk (\eg a ResNet-101 \cite{He16}) to produce convolutional features which are shared for the task of detection and tracking. We compute convolutional cross-correlation between the feature responses of adjacent frames to estimate the local displacement at different feature scales. On top of the features, we employ an RoI-pooling layer \cite{rfcnNIPS16} to classify and regress box proposals as well as an RoI-tracking layer that regresses box transformations (translation, scale, aspect ratio) across frames. Finally, to infer long-term tubes of objects across a video we link detections based on our tracklets. 

An evaluation on the large-scale ImageNet VID dataset shows that our approach is able to achieve better single-model performance than the winner of the last ILSVRC'16 challenge, despite being conceptually simple and much faster. Moreover, we show that including a tracking loss may improve feature learning for better static object detection, and we also present a very fast version of D\&T that works on temporally-strided input frames.

\section{Related work}
\label{sec:related_work}
\paragraph{Object detection.}

Two families
of detectors are currently popular: First, region proposal based
detectors R-CNN \cite{Girshick14}, Fast R-CNN \cite{Girshick15},
Faster R-CNN \cite{ren2016faster} and R-FCN \cite{rfcnNIPS16} and
second, detectors that directly predict boxes for an image in one step
such as YOLO \cite{redmon2016you} and SSD \cite{liu2016ssd}.

Our approach builds on R-FCN \cite{rfcnNIPS16} which is a simple and efficient 
framework for object detection on region proposals with a fully convolutional nature. In terms of accuracy it is competitive with Faster R-CNN \cite{ren2016faster} which uses a multi-layer network that is evaluated per-region (and thus has a cost growing linearly with the number of candidate RoIs). R-FCN reduces the cost for region classification by pushing the region-wise operations to the end of the network with the introduction of a position-sensitive RoI pooling layer which works on convolutional features that encode the spatially subsampled class scores of input RoIs.

\paragraph{Tracking.}
Tracking is also an extensively studied problem in computer vision with most recent progress devoted to trackers operating on deep ConvNet features. 
In \cite{nam2015learning} a  ConvNet is fine-tuned at test-time to track a target from the same video via detection and bounding box regression. Training on the examples of a test sequence is slow and also not applicable in the object detection setting. Other methods use pre-trained ConvNet features to track and have achieved strong performance either with correlation \cite{ma2015hierarchical,bertinetto2016fully} or regression trackers on heat maps~\cite{wang2015visual} or bounding boxes \cite{held2016learning}. The regression tracker in \cite{held2016learning} is related to our method. It is based on a Siamese ConvNet that predicts the location in the second image of the object shown in the center of the previous image. Since this tracker predicts a bounding box instead of just the position, it is able to model changes in scale and aspect of the tracked template. The major drawback of this approach is that it only can process a single target template and it also has to rely on significant data augmentation to learn all possible transformations of tracked boxes.
The approach in \cite{bertinetto2016fully} is an example of a correlation tracker and inspires our method. The tracker also uses a fully-convolutional Siamese network that takes as input the tracking template and the search image. The ConvNet features from the last convolutional layer are correlated to find the target position in the response map. One drawback of many correlation trackers \cite{ma2015hierarchical,bertinetto2016fully} is that they only work on single targets and do not account for changes in object scale and aspect ratio.

\paragraph{Video object detection.}

Action detection is also a related problem
and has received increased attention recently, mostly with methods
building on two-stream ConvNets \cite{Simonyan14b}. In
\cite{gkioxari2014finding} a method is presented that uses a
two-stream R-CNN \cite{Girshick14} to classify regions and link
them across frames based on the action predictions and their spatial
overlap. This method has been adopted by \cite{saha2016deep} and
\cite{peng2016multi} where the R-CNN was replaced by Faster R-CNN with
the RPN operating on two streams of appearance and motion information.

One area of interest is learning to detect and localize in each
frame (e.g.\ in video co-localization) 
with only weak supervision. The 
YouTube Object Dataset~\cite{prest2012learning}, has been used for this
purpose,~\eg\cite{kwak2015unsupervised,joulin2014efficient}. 

Since the object detection from video task has been introduced at the ImageNet challenge, it has drawn significant attention. In \cite{kang2016object} tubelet proposals are generated by applying a tracker to frame-based bounding box proposals. The detector scores  across the video are re-scored by a 1D CNN model. In their corresponding ILSVRC submission the group \cite{kang2016tcnn} added a propagation of scores to nearby frames based on optical flows between frames and suppression of class scores that are not among the top classes in a video. A more recent work \cite{kang2017object} introduces a tubelet proposal network that regresses static object proposals over multiple frames, extracts features by applying Faster R-CNN which are finally processed by an encoder-decoder LSTM.  In deep feature flow \cite{DFFcvpr17} a recognition ConvNet is applied to key frames only and an optical flow ConvNet is used for propagating the deep feature maps via a flow field to the rest of the frames. This approach can increase detection speed by a factor of 5 at a slight accuracy cost. The approach is error-prone due largely to two aspects: First, propagation from the key frame to the current frame can be erroneous and, second, the key frames can miss features from current frames. Very recently a new large-scale dataset for video object detection has been introduced \cite{youtubeBB} with single objects annotations over video sequences. 

\begin{figure*}[!ht]
	\centering
	\includegraphics[width=.85\linewidth]{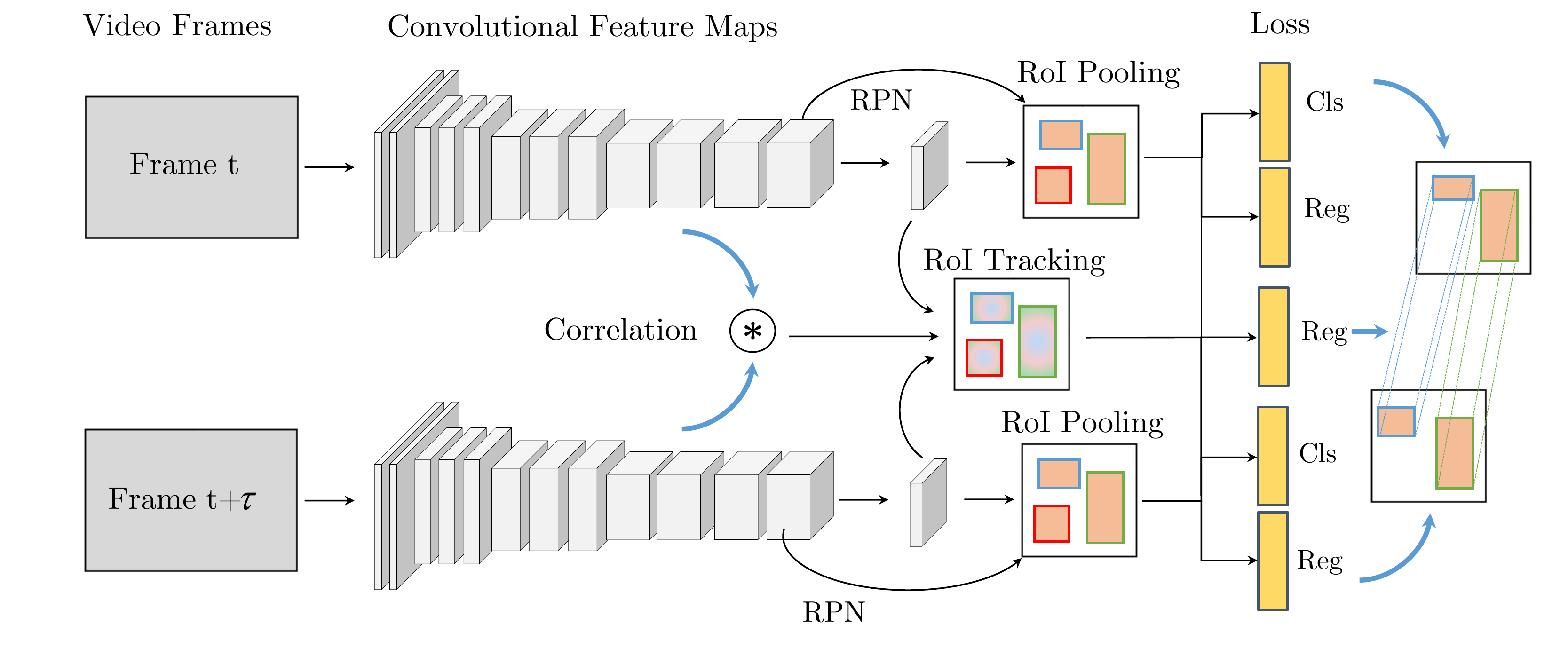}
	\caption[caption]{Architecture of our Detect and Track (D\&T) approach (see Section \protect\ref{sec:approach} for details).}\vspace{-0.15in}
	\label{fig:architecture}
\end{figure*}

\section{D\&T Approach} \label{sec:approach}

In this section we first give an overview of the Detect and Track
(D\&T) approach (\sref{sec:overview}) that generates tracklets given
two (or more) frames as input. We then give the details, starting with the baseline R-FCN
detector \cite{rfcnNIPS16} (\sref{sec:detection}),  and formulating the
tracking objective as cross-frame bounding box regression
(\sref{sec:tracking});  finally, we introduce the correlation features
(\sref{sec:correlation}) that aid the network in the tracking process.

\sref{sec:linking} shows how we link across-frame tracklets to tubes over the temporal extent of a video,
and \sref{sec:experiments} describes how we apply D\&T to the ImageNet VID challenge.

\subsection{D\&T overview} \label{sec:overview}
We aim at jointly detecting and tracking (D\&T) objects in video. \figref{fig:architecture} illustrates our D\&T architecture. We build on the R-FCN \cite{rfcnNIPS16} object detection framework which is fully convolutional up to region classification and regression, and extend it for multi-frame detection and tracking.  Given a set of two high-resolution input frames our architecture  first computes convolutional feature maps that are shared for the tasks of detection and tracking (\eg the features of a ResNet-101\cite{He16}). An RPN is used to propose candidate regions in each frame based on the objectness likelihood for pre-defined candidate boxes (\ie ``anchors''\cite{ren2016faster}). Based on these regions, RoI pooling is employed to aggregate position-sensitive score and regression maps, produced from intermediate convolutional layers, to classify boxes and refine their coordinates (regression), respectively. 

We extend this architecture by introducing a regressor that takes the intermediate position-sensitive regression maps from both frames (together with correlation maps, see below) as input to an RoI tracking operation which outputs the box transformation from one frame to the other. The correspondence between frames is thus simply accomplished by pooling features from both frames, at the same proposal region.
We train the RoI tracking task by extending the multi-task objective of R-FCN with a tracking loss that regresses object coordinates across frames. Our tracking loss operates on ground truth objects and evaluates a soft L1 norm \cite{Girshick15} between coordinates of the predicted track and the ground truth track of an object.

Such a tracking formulation can be seen as a multi-object extension of the single target tracker in \cite{held2016learning} where a ConvNet is trained to infer an object's bounding box  from features of the two frames. One drawback of such an approach is that it does not exploit translational equivariance  which means that the tracker has to learn all possible translations from training data. Thus such a tracker requires exceptional data augmentation (artificially scaling and shifting boxes) during training \cite{held2016learning} . 

A tracking representation that is based on correlation filters \cite{bolme2010visual,danelljan2016beyond,henriques2015high} can exploit the translational equivariance as correlation is equivariant to translation. Recent correlation trackers
\cite{bertinetto2016fully,ma2015hierarchical} typically work on high-level ConvNet features and compute the cross correlation between a tracking template and the search image (or a local region around the tracked position from the previous frame). The resulting correlation map measures the similarity between the template and the search image for all circular shifts along the horizontal and vertical dimension. The displacement of a target object can thus be found by taking the maximum of the correlation response map.

Different from typical correlation trackers that work on single target templates,
we aim to track multiple objects simultaneously. We compute correlation maps
for all positions in a feature map and let RoI tracking additionally operate on these feature maps for better track regression. Our architecture is able to be trained end-to-end taking as input frames from a video and producing object detections and their tracks.  The next sections describe how we structure our architecture for end-to-end learning of object detection and tracklets.

\subsection{Object detection and tracking in R-FCN} \label{sec:detection}
Our architecture takes frames $\bI^t \in \mR^{H_0 \times W_0\times 3}$ at time $t$ and pushes them through a backbone ConvNet (\ie ResNet-101 \cite{He16}) to obtain 
 feature maps $\bx^t_l \in \mathbb{R}^{H_l\times W_l\times D_l}$  where
$W_l,H_l$ and $D_l$ are the width, height and number of channels of the
respective feature map output by layer $l$. As in R-FCN \cite{rfcnNIPS16} we reduce the effective stride at the last convolutional layer from 32 pixels to 16 pixels by modifying the conv$5$ block to have unit spatial stride, and also increase its receptive field by dilated convolutions \cite{long2015fully}.

Our overall system builds on the R-FCN \cite{rfcnNIPS16} object detector which works in two stages: first it extracts candidate regions of interest (RoI) using a Region Proposal Network (RPN) \cite{ren2016faster};  and, second, it performs region classification into different object categories and background by using a position-sensitive RoI pooling layer \cite{rfcnNIPS16}. The input to this RoI pooling layer comes from an extra convolutional layer with output $\bx^t_{cls}$ that operates on the last convolutional layer of a ResNet \cite{He16}. The layer produces a bank of $D_{cls}=k^2(C+1)$ position-sensitive score maps which correspond to a $k\times k$ spatial grid describing relative positions to be used in the RoI pooling operation for each of the $C$ categories and background. Applying the softmax function to the outputs leads to a probability distribution $p$ over $C+1$ classes for each RoI. In a second branch R-FCN puts a sibling convolutional layer with output $\bx^t_{reg}$ after the last convolutional layer for bounding box regression, again a position-sensitive RoI pooling operation is performed on this bank of $D_{cls}=4 k^2$ maps for class-agnostic bounding box prediction of a box $b=(b_x, b_y, b_w, b_h)$.

\begin{figure*}[!t]
	\centering

	\includegraphics[width=.85\textwidth]{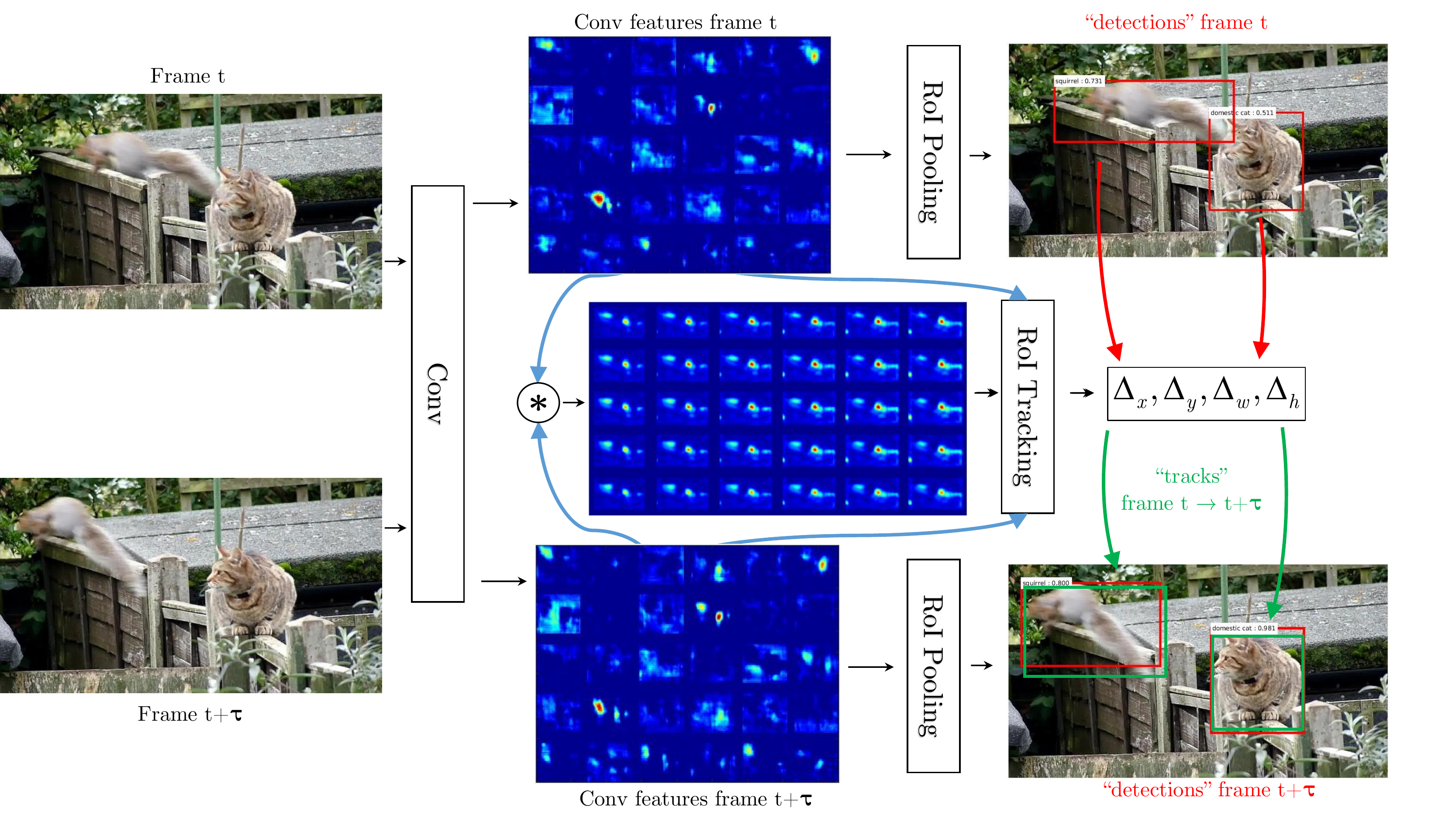}   

	\caption{Schematic of our approach for two frames at time $t$ and $t+\tau$. The
		inputs are first passed through a fully-convolutional network to produce feature
		maps. A correlation layer operates on multiple feature maps of different scales
		(only the coarsest scale is shown in the figure) and estimates local feature
		similarity for various offsets between the two frames. Finally, position
		sensitive RoI-pooling \cite{rfcnNIPS16} operates on the convolutional features
		of the individual frames to produce per-frame detections and also on a stack of
		individual frame-features as well as the between frame correlation features to output
		regression offsets of the boxes across the two frames (RoI-tracking).
	}
	\label{fig:tracking_schematic}
    \vspace{-10pt}
\end{figure*}

Let us now consider a pair of frames $\bI^{t}, \bI^{t+\tau}$, sampled at time $t$ and $t+\tau$, given as input to the network. We introduce an inter-frame bounding box regression layer that performs position sensitive RoI pooling on the concatenation of the bounding box regression features $\{\bx^{t}_{reg},\bx^{t+\tau}_{reg}\}$ to predict the transformation $\Delta^{t+\tau}=(\Delta^{t+\tau}_x, \Delta^{t+\tau}_y, \Delta^{t+\tau}_w, \Delta^{t+\tau}_h)$ of the RoIs from $t$ to $t+\tau$.  
The correlation features, that are also used by the bounding box
regressors, are described in section~\ref{sec:correlation}.
\figref{fig:tracking_schematic} shows an illustration of this approach.

\subsection{Multitask detection and tracking objective} 
\label{sec:tracking}
To learn this regressor, we extend the multi-task loss of Fast R-CNN \cite{Girshick15}, consisting of a combined classification $L_{cls}$ and regression loss $L_{reg}$, with an additional term that scores the tracking across two frames $L_{tra}$. For a single iteration and a batch of $N$ RoIs the network predicts softmax probabilities $\{p_i\}_{i=1}^N$, regression offsets $\{b_i\}_{i=1}^N$, and cross-frame RoI-tracks $\{\Delta^{t+\tau}_i\}_{i=1}^{N_{tra}}$. Our overall objective function is written as:
\begin{equation}\label{eq:loss}
\begin{aligned}
L(\{p_i\}, \{b_i\}, \{\Delta_i\}) = \frac{1}{N}  \sum_{i=1}^N L_{cls}(p_{i,c^{*}}) \\
 + \lambda \frac{1}{N_{fg}} \sum_{i=1}^N [c_i^{*}>0] L_{reg}(b_i, b_i^*)\\
+\lambda \frac{1}{N_{tra}}  \sum_{i=1}^{N_{tra}} L_{tra}(\Delta^{t+\tau}_i, \Delta^{*,t+\tau}_i).\\ 
\end{aligned}
\end{equation}
The ground truth class label of an RoI is defined by $c_i^{*}$ and its predicted softmax score is $p_{i,c^{*}}$. $b_i^*$ is the ground truth regression target, and  $\Delta^{*,t+\tau}_i$ is the track regression target.  The indicator function $[c_i^{*}>0]$
is 1 for foreground RoIs and 0 for background RoIs (with $c_i^{*}=0$). $L_{cls}(p_{i,c^{*}})=-\log(p_{i,c^{*}})$ is the cross-entropy loss for box classification, and $L_{reg}$ \& $L_{tra}$ are bounding box and track regression losses defined as the smooth L1 function in \cite{Girshick15}. The tradeoff parameter is set to $\lambda=1$ as in \cite{Girshick15,rfcnNIPS16}. 
The assignment of RoIs to ground truth is as follows: a class label $c^*$ and regression targets $b^*$ are assigned if the RoI overlaps with a ground-truth box at least by 0.5 in intersection-over-union (IoU) and the tracking target $\Delta^{*,t+\tau}$ is assigned only to ground truth targets which are appearing in both frames. Thus, the first term of \eqref{eq:loss} is active for all $N$ boxes in a training batch, the second term is active for $N_{fg}$ foreground RoIs and the last term is active for $N_{tra}$ ground truth RoIs which have a track correspondence across the two frames.

For track regression we use the bounding box regression parametrisation of R-CNN \cite{Girshick14, Girshick15, ren2016faster}. For a single object we have ground truth box coordinates  $B^t=(B^t_{x},B_{y}^{t},B_{w}^{t},B_{h}^{t})$  in frame $t$, and similarly  $B^{t+\tau}$ for frame $t+\tau$, denoting the horizontal \& vertical centre coordinates and its width and height. The tracking regression values for the target  $\Delta^{*,t+\tau}=\{\Delta^{*,t+\tau}_x,\Delta^{*,t+\tau}_y,\Delta^{*,t+\tau}_w,\Delta^{*,t+\tau}_h\}$ are then 
\begin{align}
\Delta^{*,t+\tau}_x & = \frac{B_{x}^{t+\tau} - B^t_x}{B^t_w} & \Delta^{*,t+\tau}_y & = \frac{B_{y}^{t+\tau} - B_y^t}{B_h^t}\\
\Delta^{*,t+\tau}_w & = \log(\frac{B_{w}^{t+\tau}}{B_{w}^t}) & \Delta^{*,t+\tau}_h & = \log(\frac{B_{h}^{t+\tau}}{B_{h}^t})) \text{.}
\end{align} 

\subsection{Correlation features for object tracking} \label{sec:correlation}

\begin{figure*}[!t]
	\centering
	   
	\subfigcapskip=-12pt
	\subfigtopskip=-6pt

 \resizebox {1\textwidth }{!}{ 
 	\hspace{-10pt}
	\subfigure[frame $t$]{
		\includegraphics[width=0.220\textwidth]{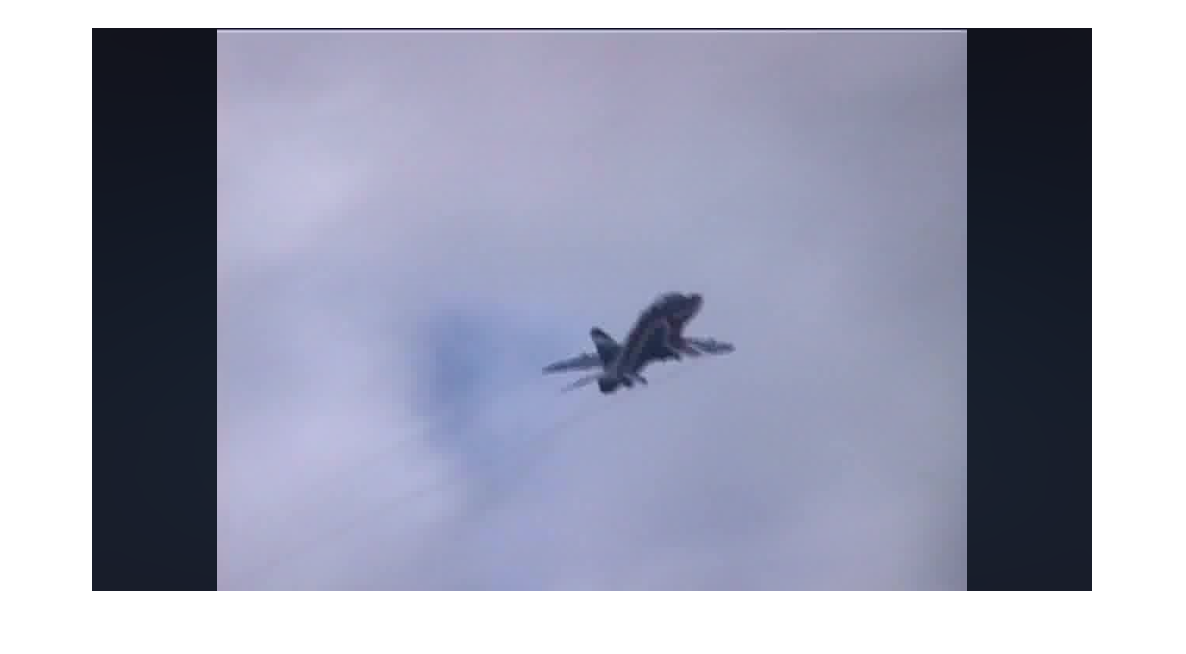}   
		\label{fig:airplane1}}\hspace{-20pt}
	\subfigure[frame $t+\tau$]{
		\includegraphics[width=0.220\textwidth]{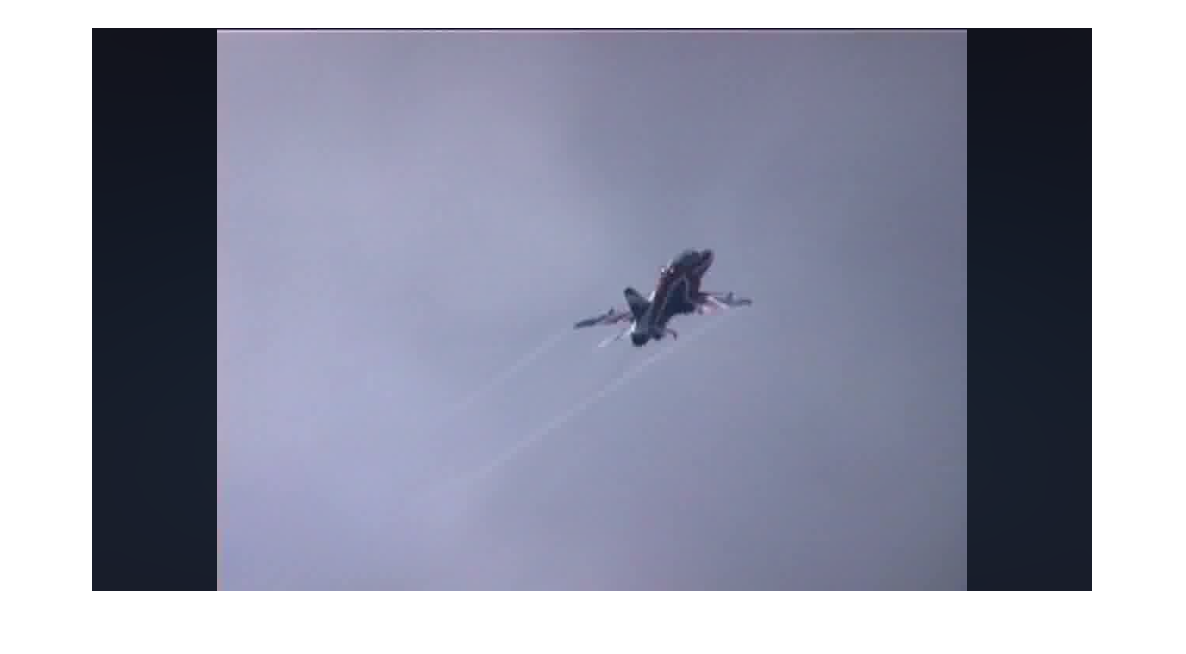}   
		\label{fig:airplane2}}\hspace{-20pt}
	\subfigure[corr.~conv3]{
		\includegraphics[width=0.230\textwidth]{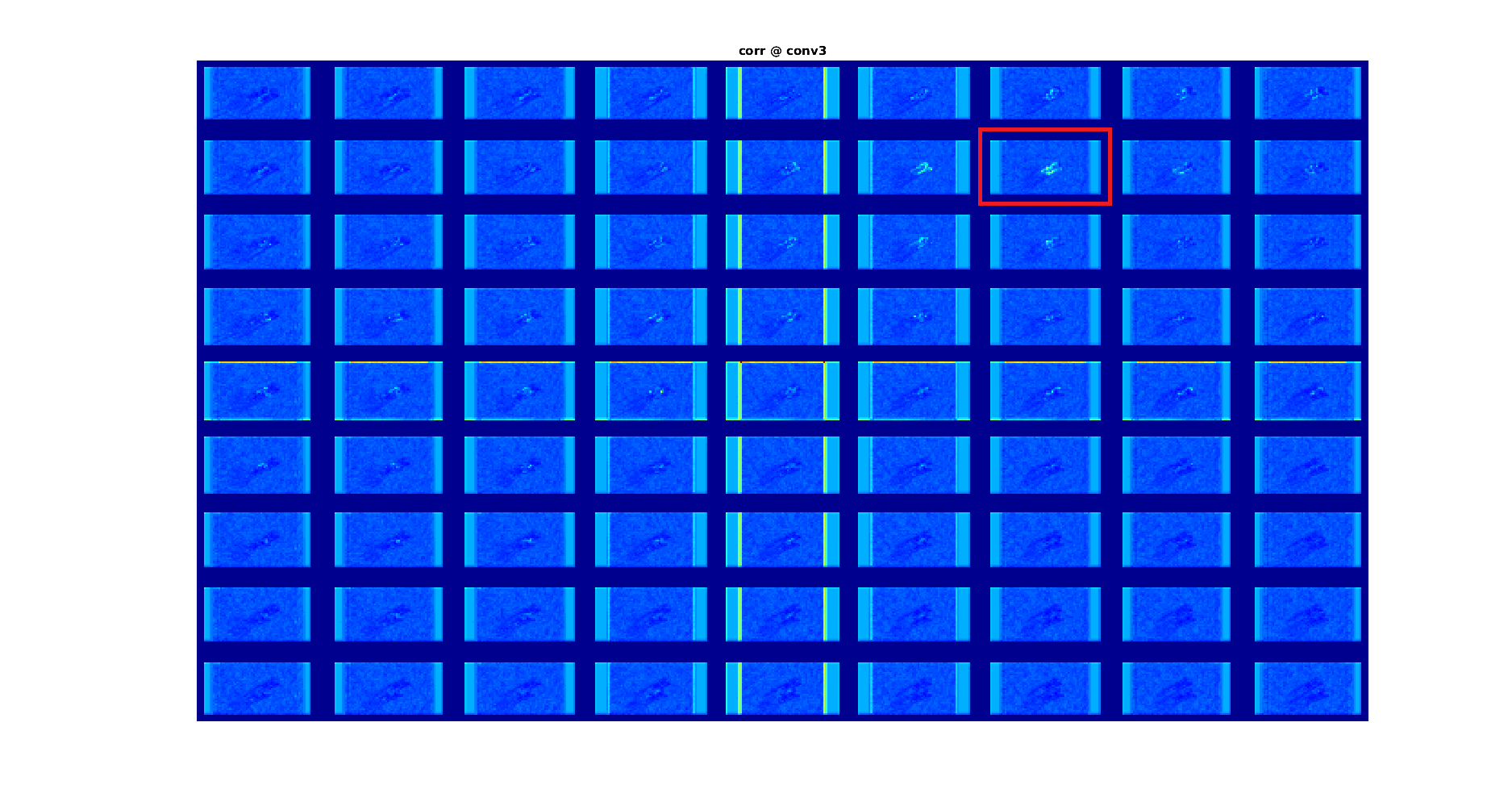}
		\label{fig:airplane_corr_conv3}}\hspace{-25pt}
	\subfigure[corr.~conv4]{
		\includegraphics[width=0.230\textwidth]{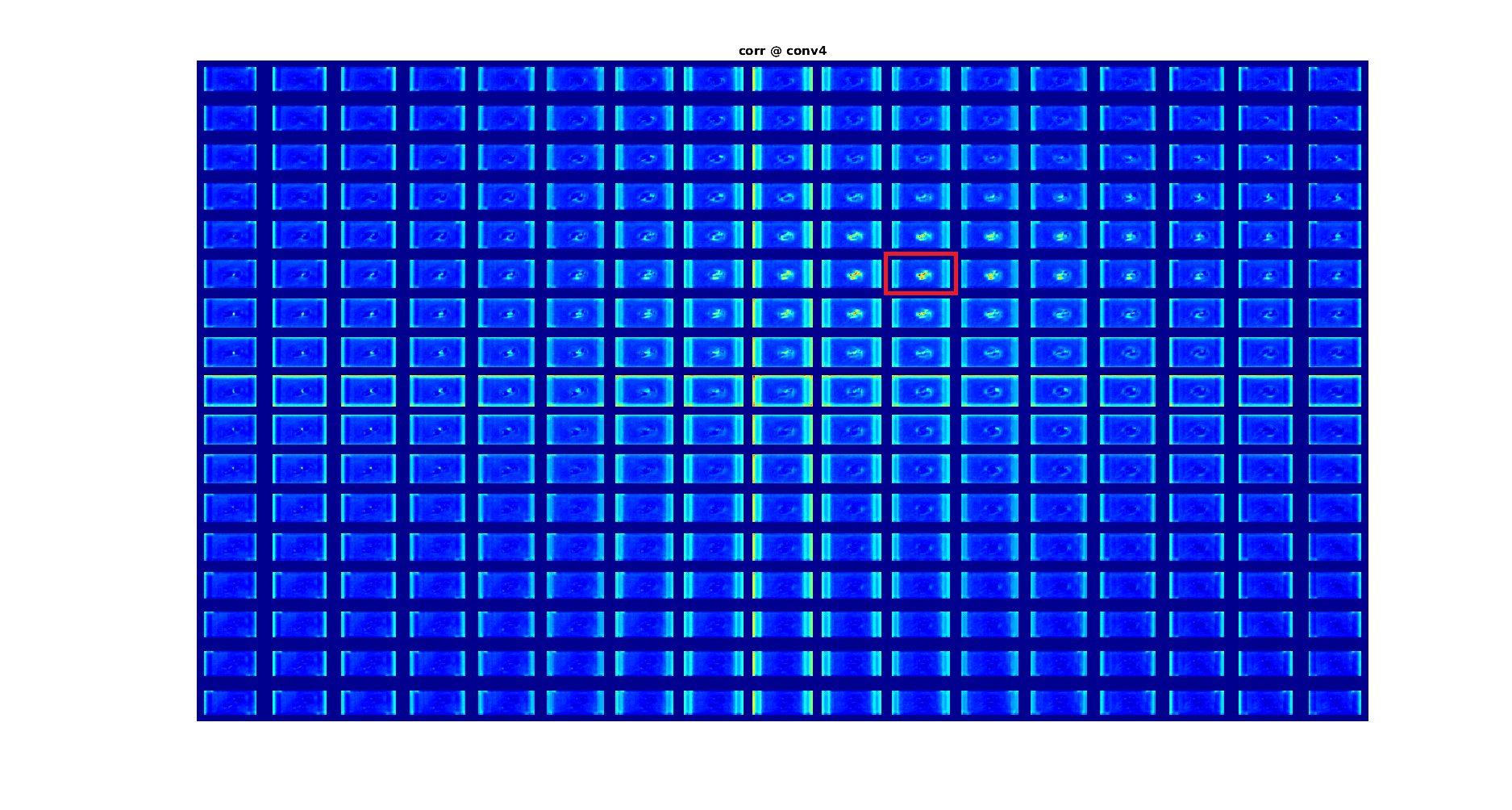}
		\label{fig:airplane_corr_conv4}}\hspace{-25pt}
	\subfigure[corr.~conv5]{
		\includegraphics[width=0.230\textwidth]{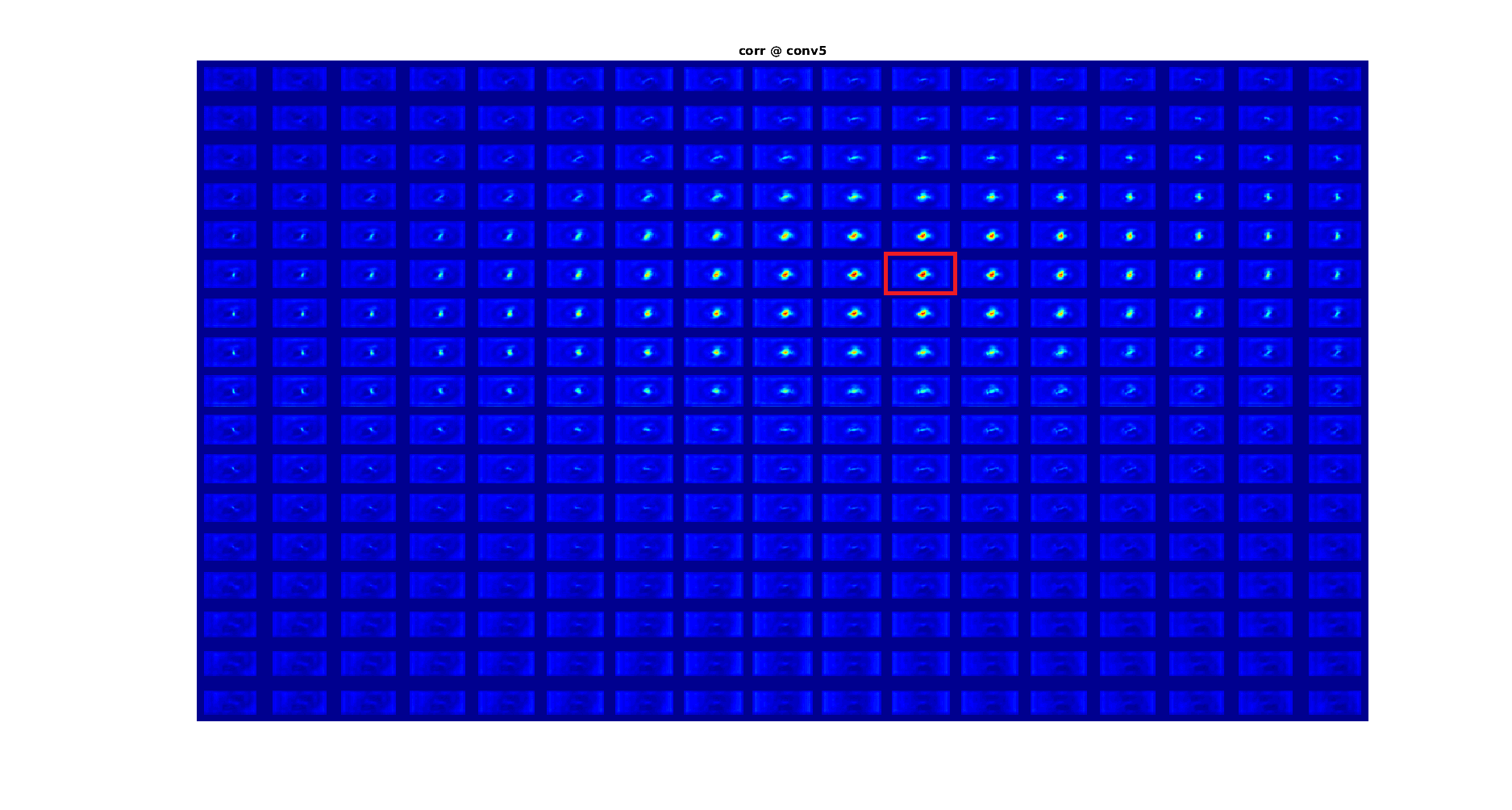}
		\label{fig:airplane_corr_conv5}}\hspace{-10pt}
 }
 \resizebox {1\textwidth }{!}{ 
 		\hspace{-10pt}
	\subfigure[frame $t$]{
		\includegraphics[width=0.220\textwidth]{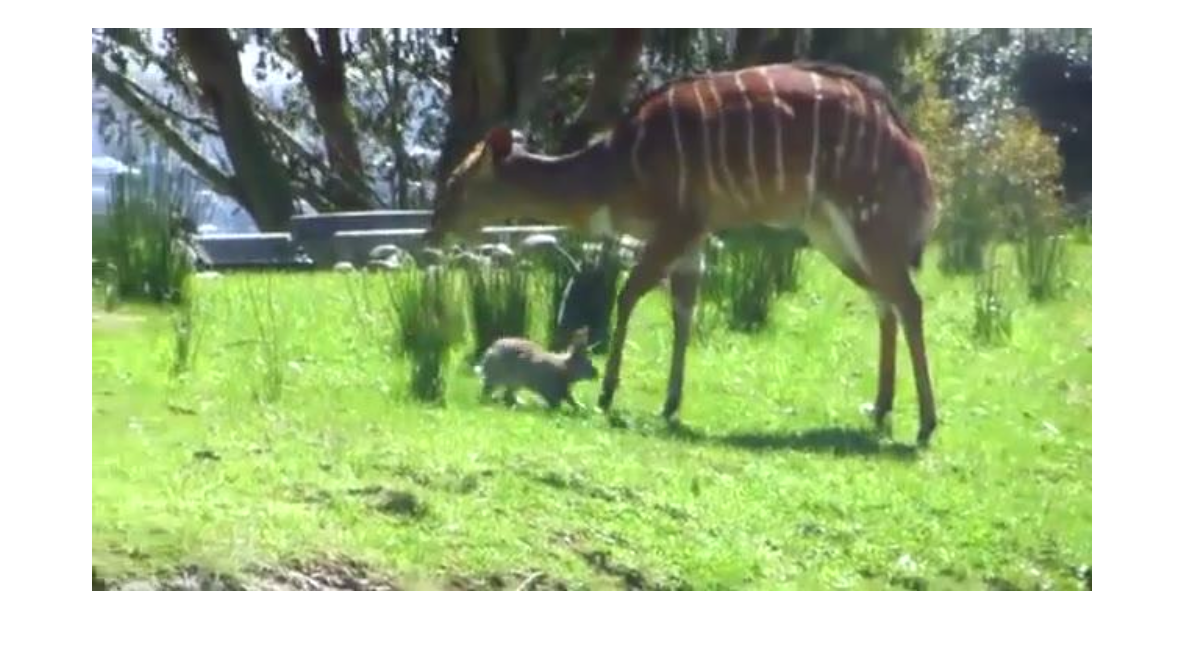}   
		\label{fig:rabbit1}}\hspace{-20pt}
	\subfigure[frame $t+\tau$]{
		\includegraphics[width=0.220\textwidth]{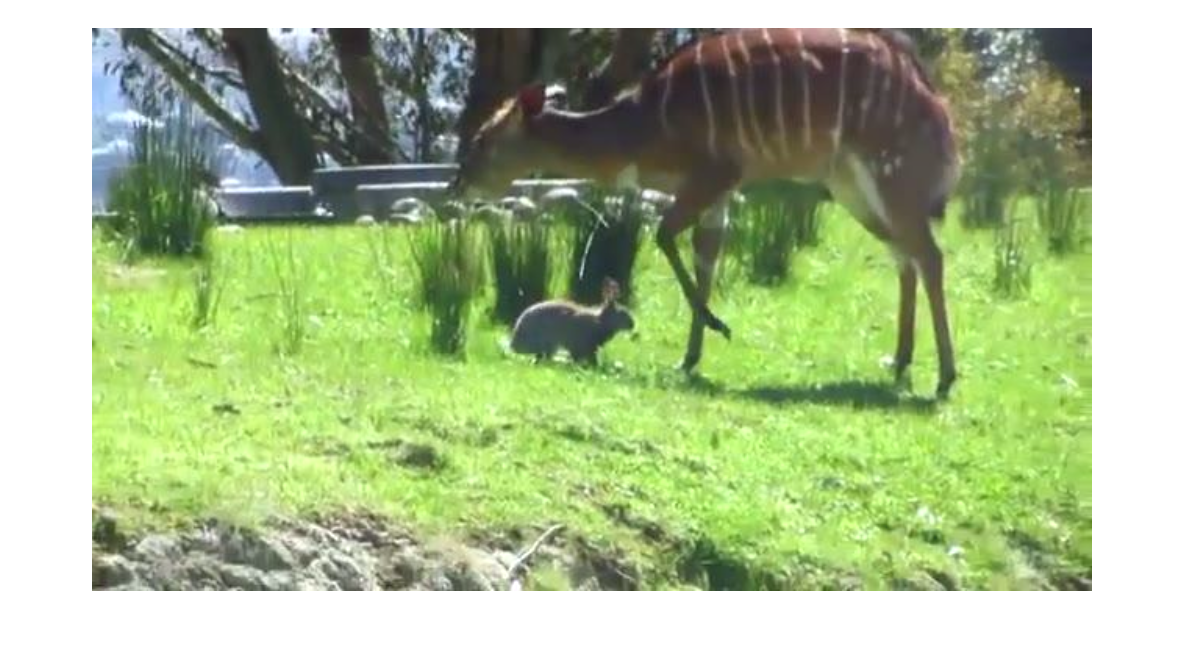}   
		\label{fig:rabbit2}}\hspace{-20pt}
	\subfigure[corr.~conv3]{
		\includegraphics[width=0.230\textwidth]{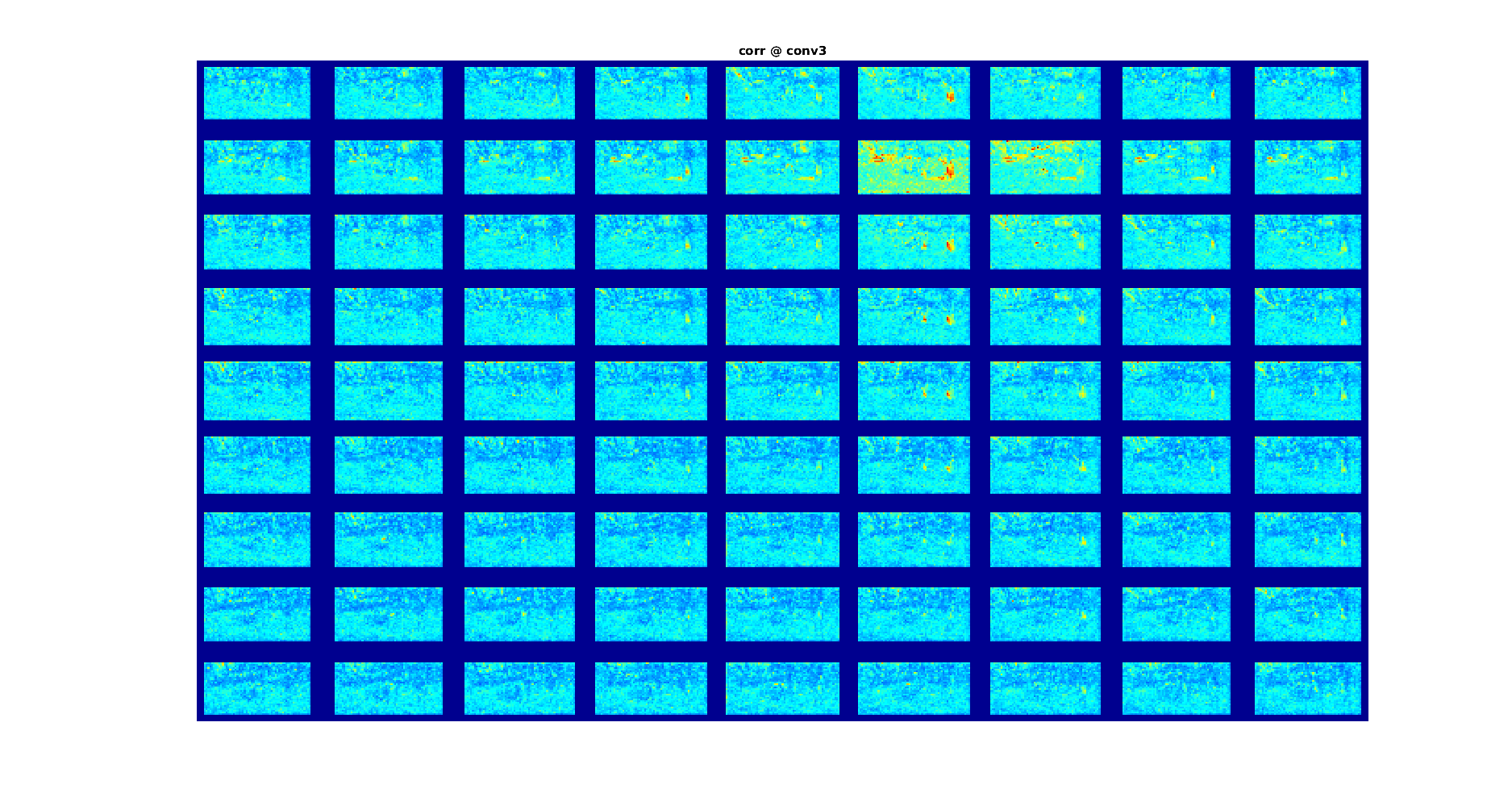}
		\label{fig:rabbit_corr_conv3}}\hspace{-25pt}
	\subfigure[corr.~conv4]{
		\includegraphics[width=0.230\textwidth]{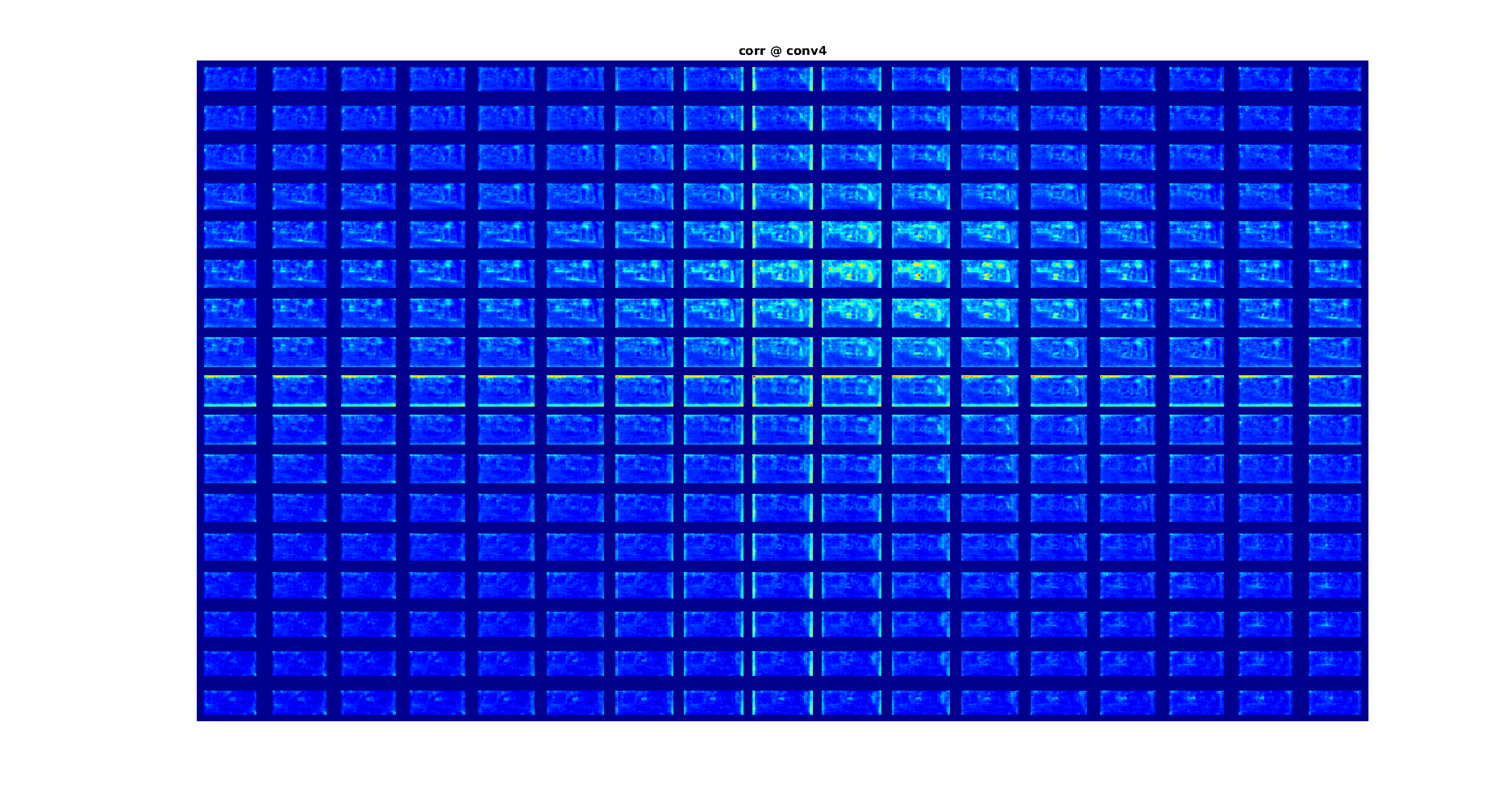}
		\label{fig:rabbit_corr_conv4}}\hspace{-25pt}
	\subfigure[corr.~conv5]{
		\includegraphics[width=0.230\textwidth]{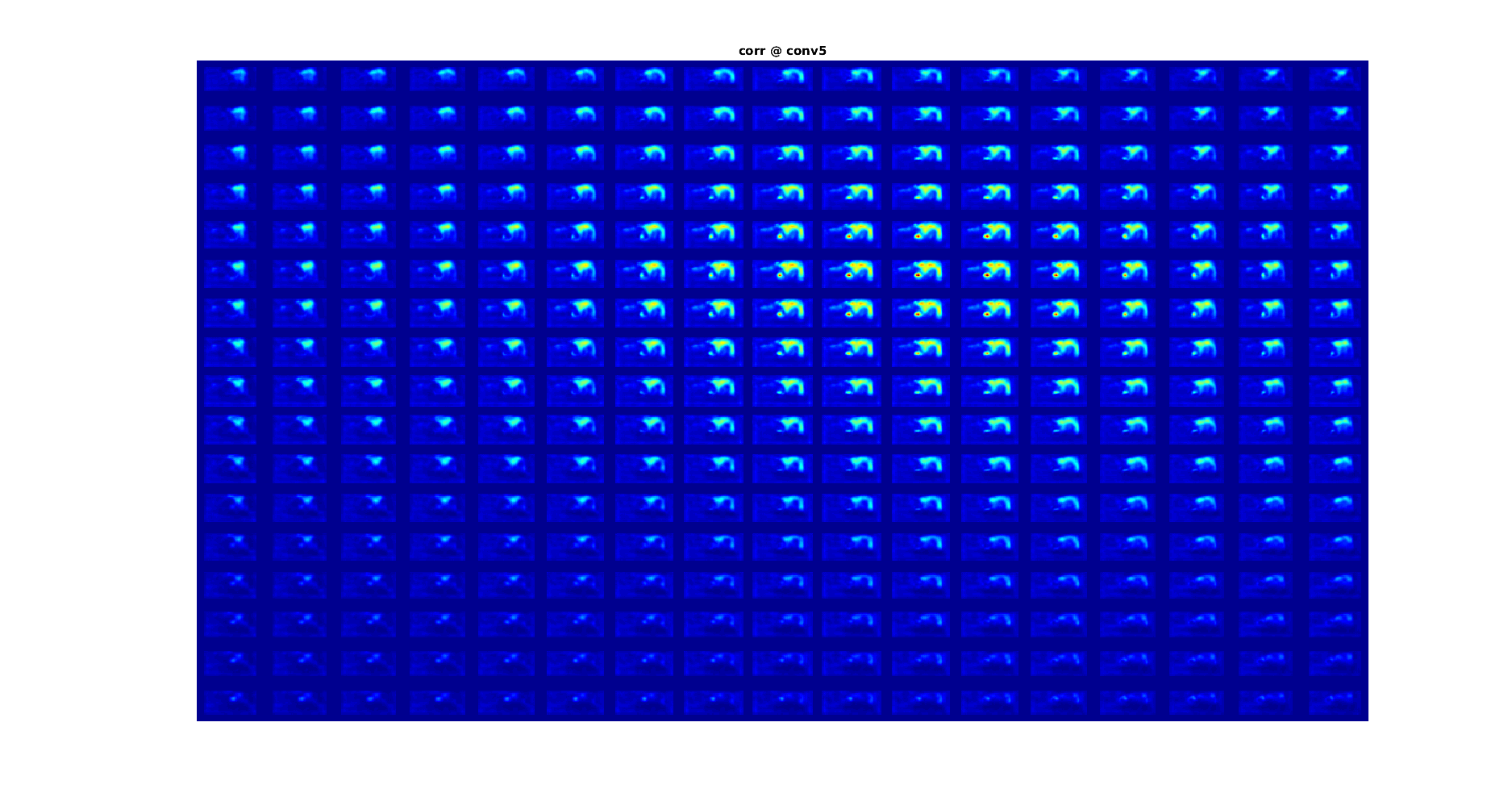}
		\label{fig:rabbit_corr_conv5}}	\hspace{-10pt}
 }
\vspace{-10pt}
	\caption{Correlation features for two frames of two validation videos. For the frames in \subref{fig:airplane1} \& \subref{fig:airplane2} , we show in \subref{fig:airplane_corr_conv3},\subref{fig:airplane_corr_conv4} and \subref{fig:airplane_corr_conv5} the correlation maps computed by using features from  conv3, conv4 and conv5, respectively. The feature maps are shown as arrays with the centre map corresponding to zero offsets $p,q$ between the frames and the neighbouring rows and columns correspond to shifted correlation maps of increasing $p,q$. We observe that the airplane moves to the top-right; hence the feature maps corresponding to $p=2,q=3$  show strong responses (highlighted in red). Note that the features at conv4 and conv5 have the same resolution, whereas at conv3 we use stride 2 correlation sampling to produce equal sized outputs. In \subref{fig:rabbit_corr_conv3},\subref{fig:rabbit_corr_conv4} and \subref{fig:rabbit_corr_conv5} we show additional multiscale correlation maps for the frames in \subref{fig:rabbit1} \& \subref{fig:rabbit2} which are affected by camera motion resulting in correlation patterns that correctly estimate this at the lower layer (conv3 corr.~responds on the grass and legs of the animal \subref{fig:rabbit_corr_conv3}), and also handles the independent motion of the animals at the higher conv5 corr~\subref{fig:rabbit_corr_conv5}.
	}

\vspace{-10pt}
	\label{fig:correlation_features}
\end{figure*}

Different from typical correlation trackers on single target templates,
we aim to track multiple objects simultaneously. We compute correlation maps
for all positions in a feature map and let RoI pooling operate on these feature
maps for track regression. Considering all possible circular shifts in a
feature map would lead to large output dimensionality and also produce responses
for too large displacements. Therefore, we restrict correlation to a local neighbourhood. This idea was originally used for optical flow estimation in
\cite{dosovitskiy2015flownet}, where a correlation layer is introduced to aid a
ConvNet in matching feature points between frames. The correlation layer
performs point-wise feature comparison of two feature maps $\bx^t_l , \bx^{t+\tau}_l $
\begin{equation}
\bx^{t,t+\tau}_{corr}(i,j,p,q)  = 
\inp[\Big]{\bx^t_l(i,j)}{\bx^{t+\tau}_l(i+p,j+q)}
\label{eq:patch_correlation}
\end{equation}
where $-d\leq p \leq d$ and $-d\leq q \leq d$ are offsets to compare features in a square neighbourhood around the locations $i,j$ in the feature map, defined by the maximum displacement, $d$. Thus the output of the correlation layer is a feature map of size $\bx_{corr} \in  \real^{H_l\times W_l\times (2d+1) \times (2d+1)}$. Equation \eqref{eq:patch_correlation} can be seen as a correlation of two feature maps within a local square window defined by $d$. We compute this local correlation for features at layers conv3, conv4 and conv5 (we use a stride of 2 in $i,j$ to have the same size in the conv3 correlation). We show an illustration of these features for two sample sequences in \figref{fig:correlation_features}. 

To use these features for track-regression, we let RoI pooling operate on these maps by stacking them with the bounding box features in \sref{sec:detection} $\{\bx^{t,t+\tau}_{corr},\bx^{t}_{reg},\bx^{t+\tau}_{reg}\}$. 

\section{Linking tracklets to object tubes} 
\label{sec:linking}
One drawback of high-accuracy object detection is that high-resolution input images have to be processed which puts a hard constraint on the number of frames a (deep) architecture can process in one iteration (due to memory limitations in GPU hardware). Therefore, a tradeoff between the number of frames and detection accuracy has to be made. Since video possesses a lot of redundant information  and objects typically move smoothly in time we can use our inter-frame tracks to link detections in time and build long-term object tubes.  To this end, we adopt an established technique from action localization \cite{gkioxari2014finding,saha2016deep,peng2016multi}, which is used to to link frame detections in time to tubes.

Consider the class detections for a frame at time $t$,  $D^{t,c}_i=\{x^t_i, y^t_i, w^t_i, h^t_i, p_{i,c}^t\}$, where $D^{t,c}_i$ is a box indexed by $i$, centred at $(x^t_i, y^t_i)$ with width $w^t_i$ and height $h^t_i$, and $p^{t}_{i,c}$ is the softmax probability for class $c$. Similarly, we also have tracks 
$T^{t,t+\tau}_i=\{x^t_i, y^t_i, w^t_i, h^t_i;   x^t_i +\Delta^{t+\tau}_x,y^t_i + \Delta^{t+\tau}_y,w^t_i + \Delta^{t+\tau}_w, h^t_i + \Delta^{t+\tau}_h  \} $ 
that describe the transformation of the boxes from frame $t$ to $t+\tau$. We can now define a class-wise linking score that combines detections and tracks across time
\begin{equation}
s_c (D_{i,c}^t,D_{j,c}^{t+\tau}, T^{t,t+\tau}) = p^{t}_{i,c} + p^{t+\tau}_{j,c}+ \psi(D_i^t,D_j,T^{t,t+\tau}) \\\\
\end{equation}
where the pairwise score is
\begin{equation}
\psi(D_{i,c}^{t},D_{j,c}^{t+\tau},T^{t,t+\tau}) = \begin{cases}
1, & \text{if } D_i^t,D_j^{t+\tau} \in T^{t,t+\tau}, \\
0, & \text{otherwise}.
\end{cases}
\end{equation}
Here, the pairwise term $\psi$ evaluates to 1 if the IoU overlap a track correspondences $T^{t,t+\tau}$ with the detection boxes $D_i^t,D_j^{t+\tau} $ is larger than 0.5. This is necessary, because the output of the track regressor does not have to exactly match the output of the box regressor. 

The optimal path across a video can then be found by maximizing the scores over the duration $\mathcal{T}$ of the video \cite{gkioxari2014finding}
\begin{equation}
\bar{D}^\star_c= \argmax_{\bar{D}}  \frac{1}{\mathcal{T}} \sum_{t=1}^{\mathcal{T}-\tau} s_c (D^t,D^{t+\tau}, T^{t,t+\tau}) 
\label{eq:viterbi}.
\end{equation}
Eq.~\eqref{eq:viterbi} can be solved efficiently by applying the Viterbi algorithm \cite{gkioxari2014finding}. Once the optimal tube $\bar{D}^\star_c$ is found, the detections corresponding to that tube are removed from the set of regions and  \eqref{eq:viterbi} is applied again to the remaining regions. 

After having found the class-specific tubes $\bar{D}_c$ for one video, we re-weight all detection scores in a tube by adding the mean of the $\alpha=50\%$ highest scores in that tube. We found that overall performance is largely robust to that parameter, with less than 0.5\% mAP variation when varying  $10\%\leq \alpha \leq 100\%$. Our simple tube-based re-weighting aims to boost the scores for positive boxes on which the detector fails. 
Using the highest scores of a tube for reweighting acts as a form of non-maximum suppression. It is also inspired by the hysteresis tracking in the Canny edge detector. Our reweighting assumes that the detector fails at most in half of a tube’s frames, and improves robustness of the tracker, though the performance is quite insensitive to the proportion chosen ($\alpha$). Note that our approach enforces the tube to span the whole video and, for simplicity, we do not prune any detections in time. Removing detections with subsequent low scores along a tube (\eg \cite{peng2016multi,saha2016deep}) could clearly improve the results, but we leave that for future work. In the following section our approach is applied to the video object detection task.

\section{Experiments}
\label{sec:experiments}

\subsection{Dataset sampling and evaluation} 

We evaluate our method on the ImageNet \cite{ILSVRC15} object detection from video (VID) dataset\footnote{\url{http://www.image-net.org/challenges/LSVRC/}} which contains 30 classes in 3862 training and 555 validation videos. The objects have ground truth annotations of their bounding box and track ID in a video. Since the ground truth for the test set is not publicly available, we measure performance as mean average precision (mAP) over the 30 classes on the validation set by following the protocols in \cite{kang2016tcnn,kang2016object,kang2017object,DFFcvpr17}, as is standard practice.

The 30 object categories in ImageNet VID are a subset of the 200 categories in the ImageNet DET dataset. Thus we follow previous approaches \cite{kang2016tcnn,kang2016object,kang2017object,DFFcvpr17} and train our R-FCN detector on an intersection of ImageNet VID and DET set (only using the data from the 30 VID classes). Since the DET set contains large variations in the number of samples per class, we sample at most 2k images per class from DET. We also subsample the VID training set by using only 10 frames from each video. The subsampling reduces the effect of dominant classes in DET (\eg there are 56K images for the dog class in the DET training set) and very long video sequences in the VID training set. 

\subsection{Training and testing}

\noindent\textbf{RPN.} Our RPN is trained as originally proposed \cite{ren2016faster}. We attach two sibling convolutional layers to the stride-reduced ResNet-101 (\sref{sec:detection}) to perform proposal classification and bounding box regression at 15 anchors corresponding to 5 scales and 3 aspect ratios. As in \cite{ren2016faster} we also extract proposals from 5 scales and apply non-maximum suppression (NMS) with an IoU threshold of 0.7 to select the top 300 proposals in each frame for training/testing our R-FCN detector. We found that pre-training on the full ImageNet DET set helps to increase the recall; thus, our RPN is first pre-trained on the 200 classes of ImageNet DET before fine-tuning on only the 30 classes which intersect ImageNet DET and VID. Our 300 proposals per image achieve a mean recall of 96.5\% on the ImageNet VID validation set. 

\noindent\textbf{R-FCN.} 
Our R-FCN detector is trained similar to \cite{rfcnNIPS16,DFFcvpr17}. We use the stride-reduced ResNet-101 with dilated convolution in conv5 (see \sref{sec:detection}) and online hard example mining \cite{ohem_cvpr16}. A randomly initialized $3\times3$, dilation 6 convolutional layer is attached to conv5 for reducing the feature dimension to 512 \cite{DFFcvpr17} (in the original R-FCN this is a $1\times1$ convolutional layer without dilation and an output dimension of 1024). For object detection and box regression, two sibling $1\times1$ convolutional layers provide the $D_{cls}=k^2(C+1)$ and $D_{reg}=4 k^2$ inputs to the position-sensitive RoI pooling layer. We use a $k\times k=7\times 7$ spatial grid for encoding relative positions as in \cite{rfcnNIPS16}. 

In both training and testing, we use single scale images with shorter dimension of 600 pixels. We use a batch size of 4 in SGD training and a learning rate of $10^{-3}$ for 60K iterations followed by  a learning rate of $10^{-4}$  for 20K  iterations. For testing we apply NMS with IoU threshold of 0.3.

\noindent\textbf{D \& T.} 
For training our D\&T architecture we start with the R-FCN model from
above and further fine-tune it on the full ImageNet VID training set
with randomly sampling a set of two adjacent frames from a different
video in each iteration. In each other iteration we also sample from
the ImageNet DET training set to avoid biasing our model to the VID
training set. When sampling from the DET set we send the same two
frames through the network as there are no sequences
available. Besides not forgetting the images from the DET training
set, this has an additional beneficial effect of letting our model
prefer small motions over large ones (\eg the tracker in
\cite{held2016learning} samples motion augmentation from a Laplacian distribution
with zero mean to bias a regression tracker on small displacements). Our
correlation features \eqref{eq:patch_correlation} are computed at
layers conv3, conv4 and conv5 with a maximum displacement of $d=8$ and
a stride of 2 in $i,j$ for the the conv3 correlation. 
For training, we use a learning rate of $10^{-4}$ for 40K iterations and $10^{-5}$ for 20K iterations at a batch size of 4.
During testing
our architecture is applied to a sequence with temporal stride $\tau$,
predicting detections $D$ and tracklets $T$ between them. For object-centred tracks, we use the regressed frame boxes as input of the ROI-tracking layer. We perform
non-maximum suppression with bounding-box voting
\cite{gidaris2015object} before the tracklet linking step to reduce the number of detections per image and
class to 25. These detections are then used in eq. \eqref{eq:viterbi} to
extract tubes and the corresponding detection boxes are re-weighted as outlined in \sref{sec:linking} for evaluation.

\begin{table*}[t]
	\centering
	\scriptsize{
		\begin{tabular}{p{2.45cm}|p{0.5cm}p{0.5cm}p{0.5cm}p{0.5cm}p{0.5cm}p{0.5cm}p{0.5cm}p{0.5cm}p{0.5cm}p{0.5cm}p{0.5cm}p{0.5cm}p{0.5cm}p{0.5cm}p{0.5cm}p{0.5cm}}
			\hline
			Methods & \rotatebox{60}{airplane} &  \rotatebox{60}{antelope} &  \rotatebox{60}{bear} &  \rotatebox{60}{bicycle} &  \rotatebox{60}{bird} &  \rotatebox{60}{bus} &  \rotatebox{60}{car} &  \rotatebox{60}{cattle} &  \rotatebox{60}{dog} &  \rotatebox{60}{d.~cat} &  \rotatebox{60}{elephant} &  \rotatebox{60}{fox} &  \rotatebox{60}{g.~panda} &  \rotatebox{60}{hamster} &  \rotatebox{60}{horse} &  \rotatebox{60}{lion} \\\hline
			TCN \cite{kang2016object} & {72.7} &  {75.5} &  42.2 &  {39.5} &  {25.0} &  64.1 &  36.3 &  {51.1} &  {24.4} &  48.6 &  65.6 &  {73.9} &  {61.7} &  {82.4} &  {30.8} &  {34.4} \\
			TPN+LSTM \cite{kang2017object}& {84.6} & 78.1 & {72.0} & {67.2} & 68.0 & 80.1 & {54.7} & 61.2 & {61.6} & {78.9} & {71.6} & 83.2 & {78.1} & 91.5 & {66.8} & {21.6}\\
			Winner ILSVRC'15 \cite{kang2016tcnn} & 83.7 &  85.7 &  84.4 &  74.5 &  73.8 &  75.7 &  57.1 &  58.7 &  72.3 &  69.2 &  80.2 &  83.4 &  80.5 &  93.1 &  84.2 &  67.8 \\
			\hline
			D (R-FCN) & 87.4&79.4&84.5&67.0&72.1&84.6&54.6&72.9&70.9&77.3&76.7&89.7&77.6&88.5&74.8&57.9\\
			D (\& T loss) & 89.4  &80.4     &    83.8    &     70.0     &    71.8     &   82.6        & 56.8    &     71.0     &   71.8     &    76.6   &      79.3    &     89.9   &      83.3  &91.9    &     76.8  &       57.3 \\
	
			D\&T ($\tau=10$)&89.1 & 79.8 & 87.5 & 68.8 & 72.9 & 86.1 & 55.7 & 78.6 & 76.4 & 83.4 & 82.9 & 97.0 & 85.0 & 96.0 & 82.2 & 66.0 \\
			D\&T ($\tau=1$) & 90.2 & 82.3 & 87.9 & 70.1 & 73.2 & 87.7 & 57.0 & 80.6 & 77.3 & 82.6 & 83.0 & 97.8 & 85.8 & 96.6 & 82.1 & 66.7   \\

			D\&T (Inception-v4)& 91.5 & 81.4 & 91.4 & 76.4 & 73.9 & 86.8 & 57.8 & 81.8 & 83.0 & 92.0 & 81.3 & 96.6 & 84.3 & 98.0 & 83.8 & 78.1 \\

			\hline
			Methods & \rotatebox{60}{lizard} &  \rotatebox{60}{monkey} &  \rotatebox{60}{motorcycle} &  \rotatebox{60}{rabbit} &  \rotatebox{60}{red panda} &  \rotatebox{60}{sheep} &  \rotatebox{60}{snake} &  \rotatebox{60}{squirrel} &  \rotatebox{60}{tiger} &  \rotatebox{60}{train} &  \rotatebox{60}{turtle} &  \rotatebox{60}{watercraft} &  \rotatebox{60}{whale} &  \rotatebox{60}{zebra} &  \multicolumn{2}{|c}{\rotatebox{60}{mAP (\%)} }  \\\hline
			TCN \cite{kang2016object} & 54.2 &  1.6 &  {61.0} &  {36.6} &  19.7 &  55.0 &  38.9 &  {2.6} &  {42.8} &  54.6 &  66.1 &  {69.2} &  {26.5} &  {68.6} &   \multicolumn{2}{|c}{47.5 }  \\
			TPN+LSTM \cite{kang2017object} & 74.4 & {36.6} & 76.3 & 51.4 & 70.6 & {64.2} & 61.2 & {42.3} & {84.8} & 78.1 & 77.2 & {61.5} & {66.9} & 88.5 & \multicolumn{2}{|c}{68.4}\\
			Winner ILSVRC'15 \cite{kang2016tcnn} & 80.3 &  54.8 &  80.6 &  63.7 &  85.7 &  60.5 &  72.9 &  52.7 &  89.7 &  81.3 &  73.7 &  69.5 &  33.5 &  90.2 &  \multicolumn{2}{|c}{73.8} \\
			Winner ILSVRC'16 \cite{ilsvrc16NUIST}  & \multicolumn{4}{c}{(single model performance)} &&&&&&&&&&& \multicolumn{2}{|c}{76.2} \\
			\hline
			D (R-FCN) &76.8&50.1&80.2&61.3&79.5&51.9&69.0&57.4&90.2&83.3&81.4&68.7&68.4&90.9&\multicolumn{2}{|c}{74.2}\\
			D (\& T loss) &     79.0       &   54.1      &    80.3    &      65.3   &      85.3      &    56.9     &     74.1    &      59.9      &    91.3    &      84.9      &        81.9     &     68.3      &    68.9      &    90.9& \multicolumn{2}{|c}{75.8}\\

			D\&T ($\tau=10$)  & 83.1 & 57.9 & 79.8 & 72.7 & 90.0 & 59.4 & 75.6 & 65.4 & 90.5 & 85.6 & 83.3 & 68.3 & 66.5 & 93.2  &\multicolumn{2}{|c}{78.6}\\
			
			D\&T ($\tau=1$) & 83.4 & 57.6 & 86.7 & 74.2 & 91.6 & 59.7 & 76.4 & 68.4 & 92.6 & 86.1 & 84.3 & 69.7 & 66.3 & 95.2 &             
			\multicolumn{2}{|c}{79.8}\\

			D\&T (Inception-v4)   & 87.9  & 59.4 & 89.3 & 78.6 & 94.7 & 63.3 & 80.1 & 71.9 & 92.5 & 84.6 & 85.5 & 70.1 & 67.4 & 95.2 & \multicolumn{2}{|c}{\textbf{82.0}}\\

			\hline
		\end{tabular}
	}
	    \vspace{-5pt}
	\caption{Performance comparison on the ImageNet VID validation set. The average precision (in \%) for each class and the mean average precision over all classes is shown. $\tau$ corresponds to the temporal sampling stride. Our D\&T variants use ResNet-101 \cite{He16} as backbone, except for the last row which lists performance for an Inception-v4 \cite{szegedy2017inception} backbone which excels at some challenging object categories. }
	\label{tab:vid_val}
    	    \vspace{-10pt}
\end{table*}

\subsection{Results}
\begin{figure*}[h]
	\centering
	
	\subfigcapskip=-12pt
	\subfigtopskip=-6pt
	
	\resizebox {1.0\textwidth }{!}{ 
		\includegraphics[width=0.11\textwidth]{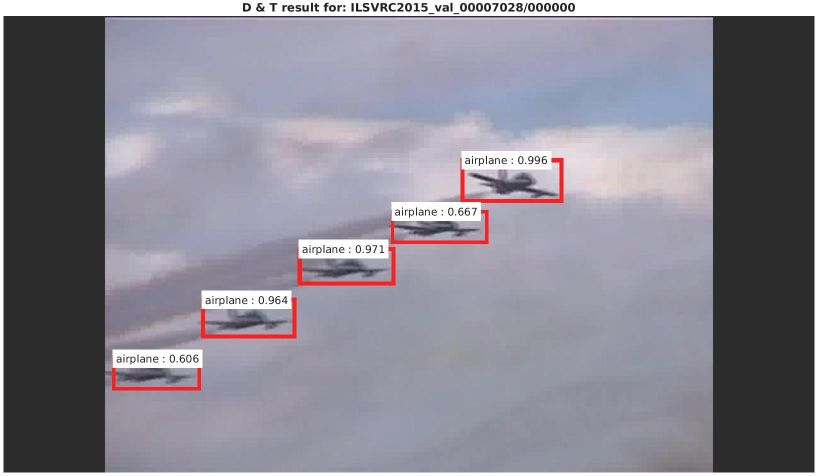}  
		\includegraphics[width=0.11\textwidth]{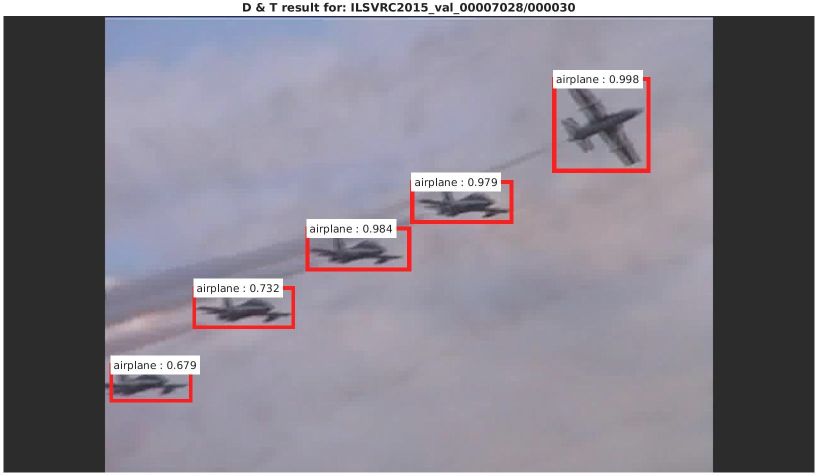}
		\includegraphics[width=0.11\textwidth]{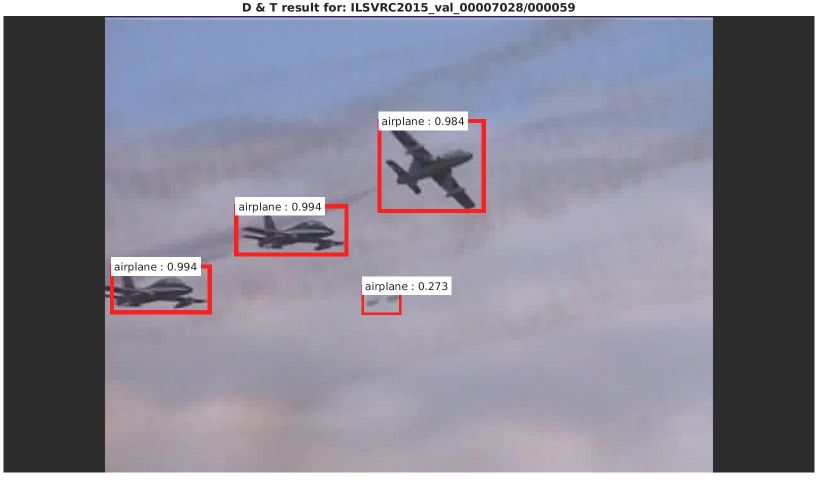}
		\includegraphics[width=0.11\textwidth]{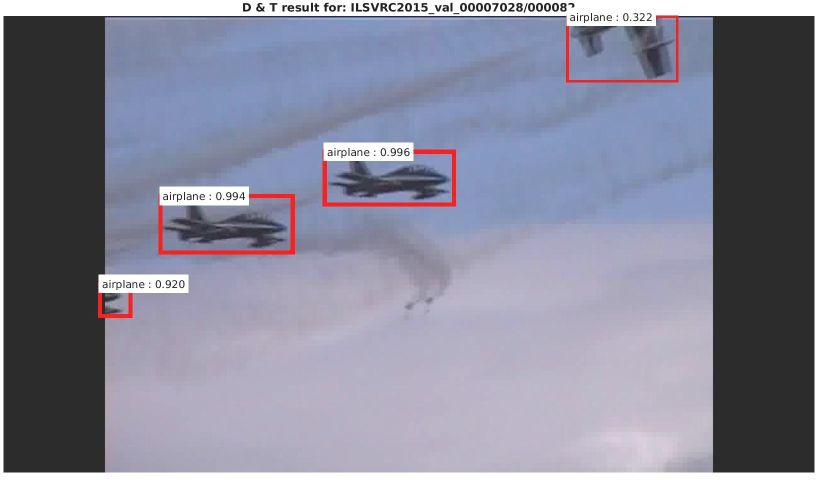}
	}
	
	\resizebox {1.0\textwidth }{!}{ 
		\includegraphics[width=0.11\textwidth]{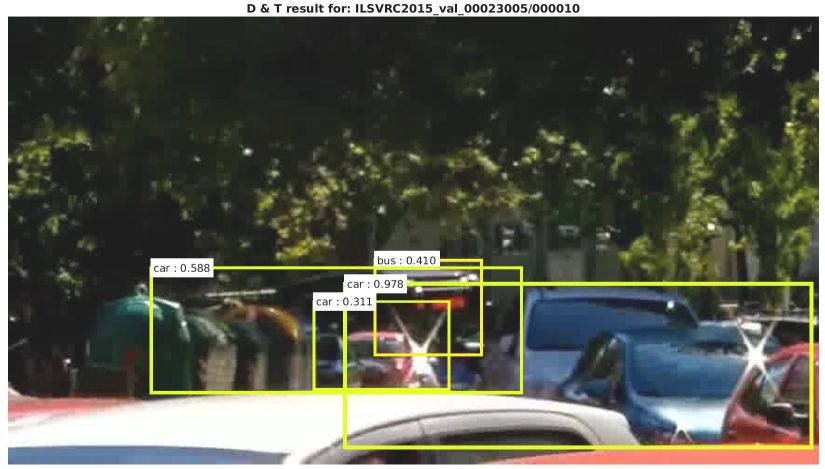}  
		\includegraphics[width=0.11\textwidth]{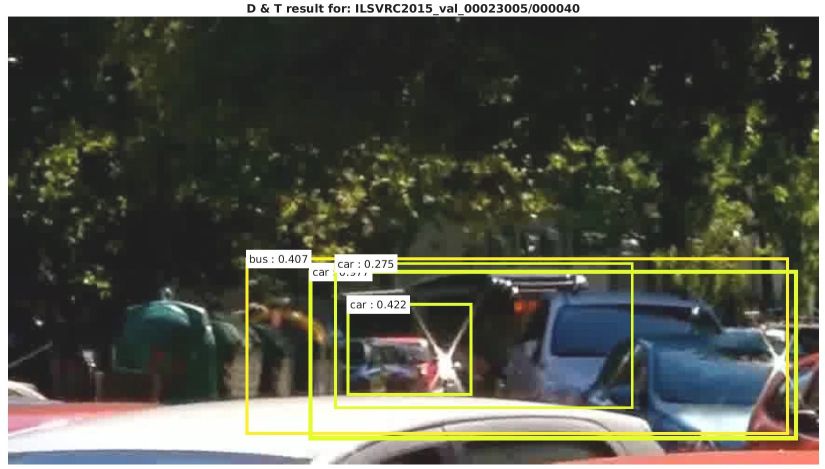}
		\includegraphics[width=0.11\textwidth]{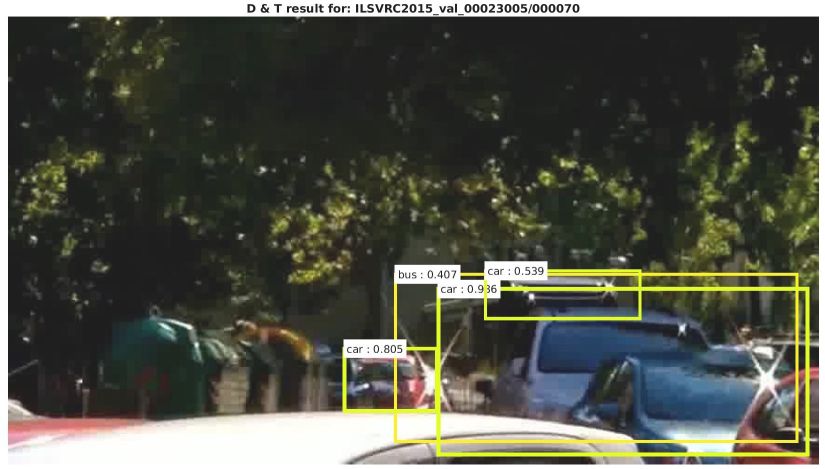}
		\includegraphics[width=0.11\textwidth]{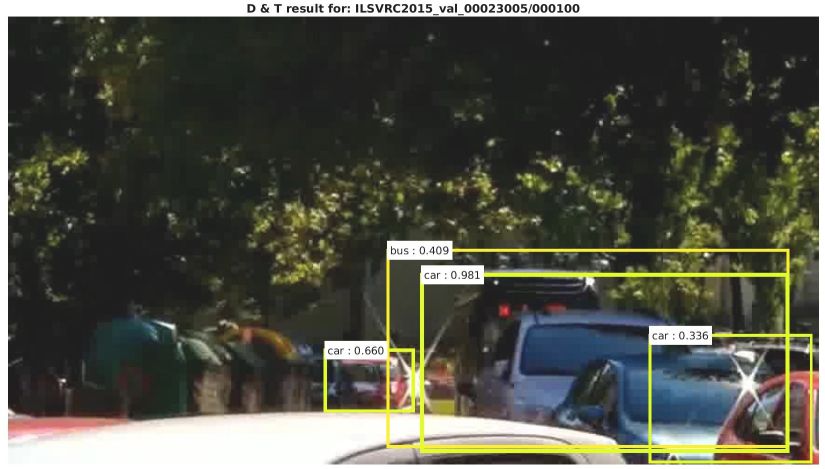}
	}	
	
	\resizebox {1.0\textwidth }{!}{ 
		\includegraphics[width=0.11\textwidth]{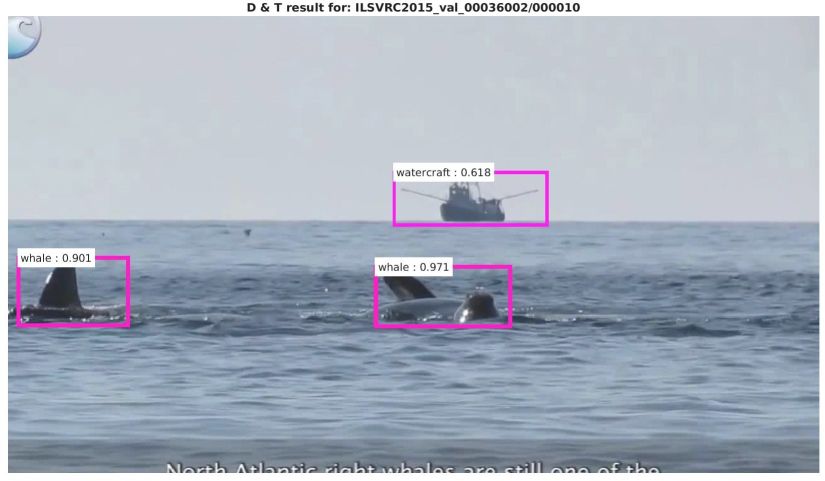} 
		\includegraphics[width=0.11\textwidth]{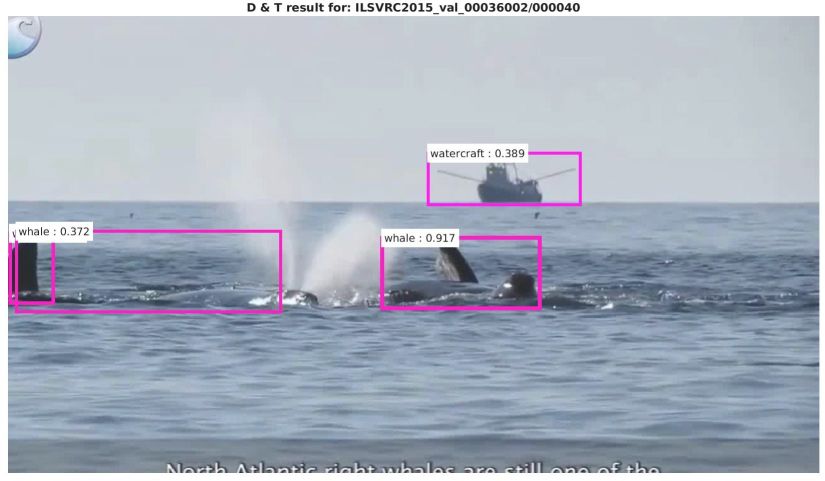}
		\includegraphics[width=0.11\textwidth]{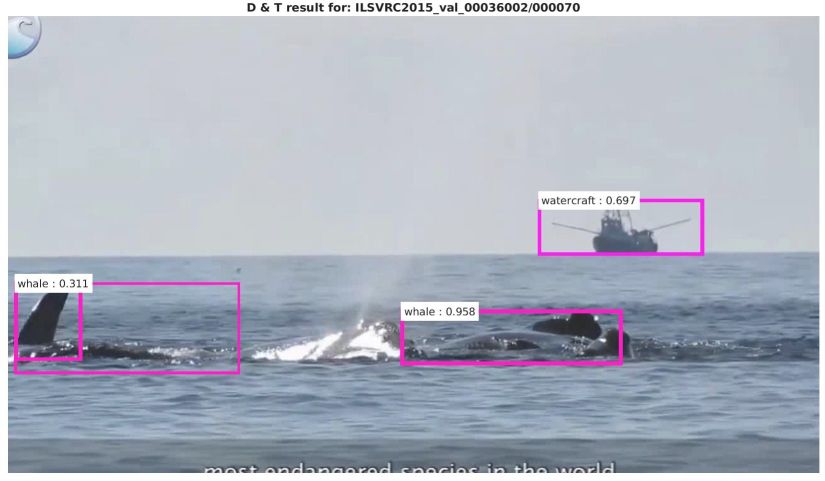}
		\includegraphics[width=0.11\textwidth]{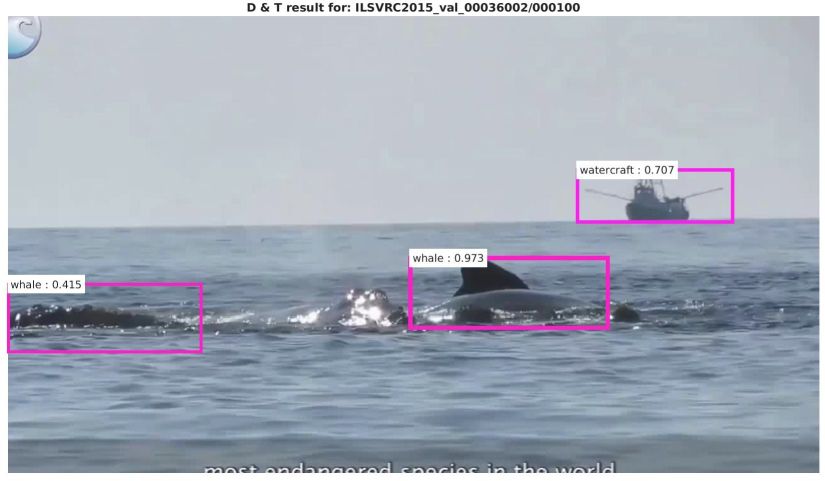}
	}
	
	\resizebox {1.0\textwidth }{!}{ 
		\includegraphics[width=0.11\textwidth]{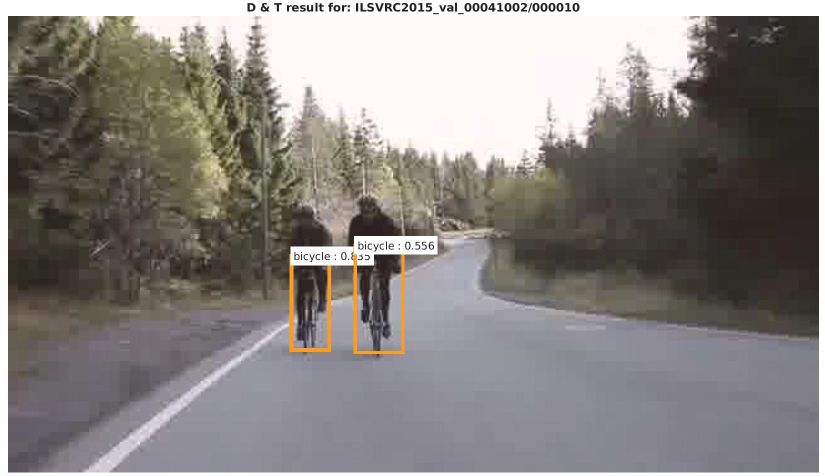} 
		\includegraphics[width=0.11\textwidth]{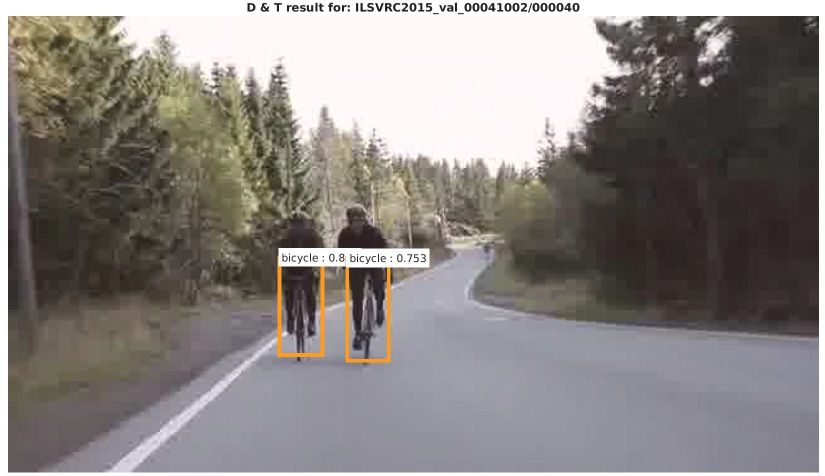}
		\includegraphics[width=0.11\textwidth]{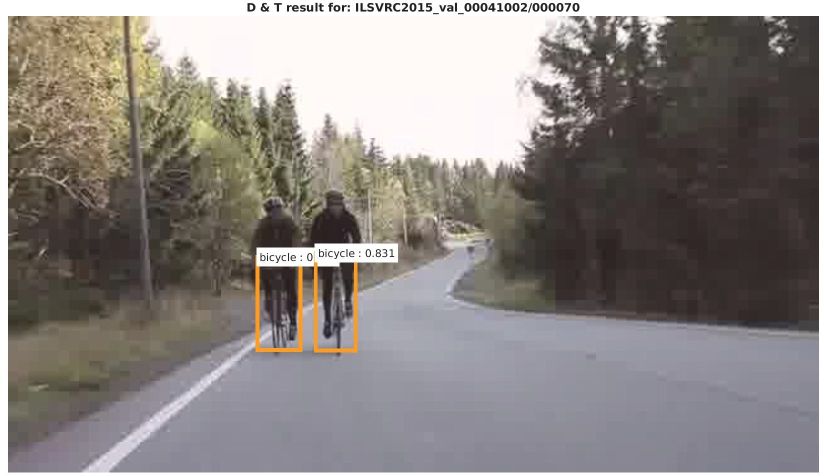}
		\includegraphics[width=0.11\textwidth]{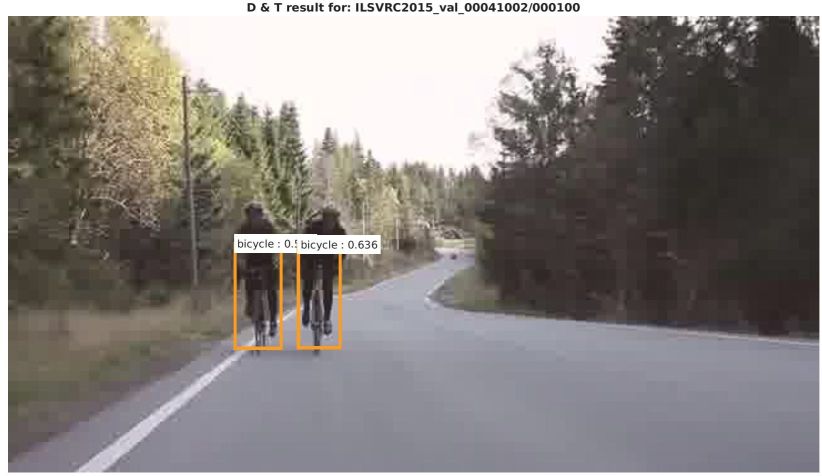}
	}
	
	\resizebox {1.0\textwidth }{!}{ 
		\includegraphics[width=0.11\textwidth]{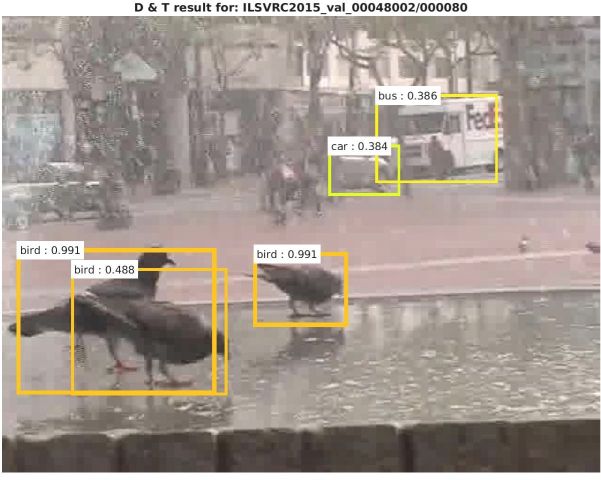}
		\includegraphics[width=0.11\textwidth]{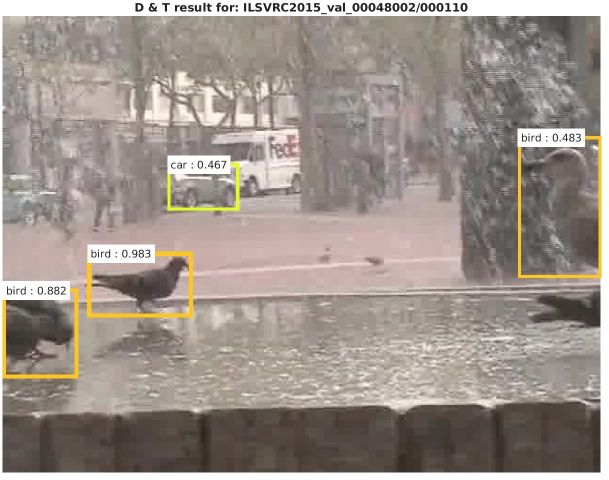}
		\includegraphics[width=0.11\textwidth]{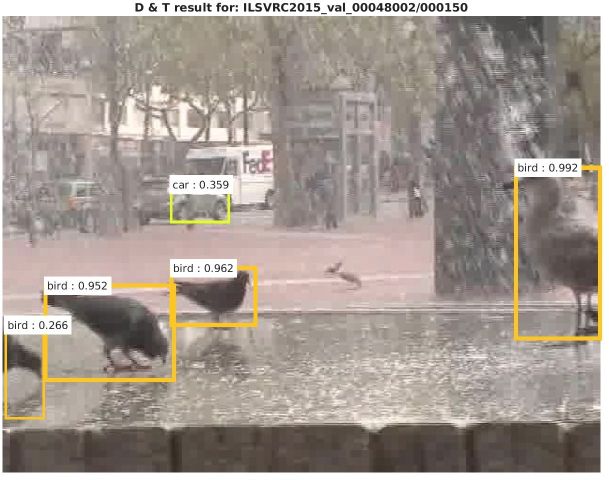}
		\includegraphics[width=0.11\textwidth]{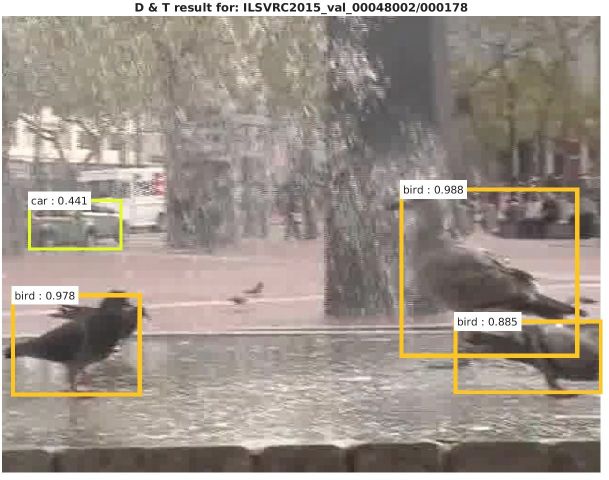}
	}

	\resizebox {1.0\textwidth }{!}{ 
		\includegraphics[width=0.11\textwidth]{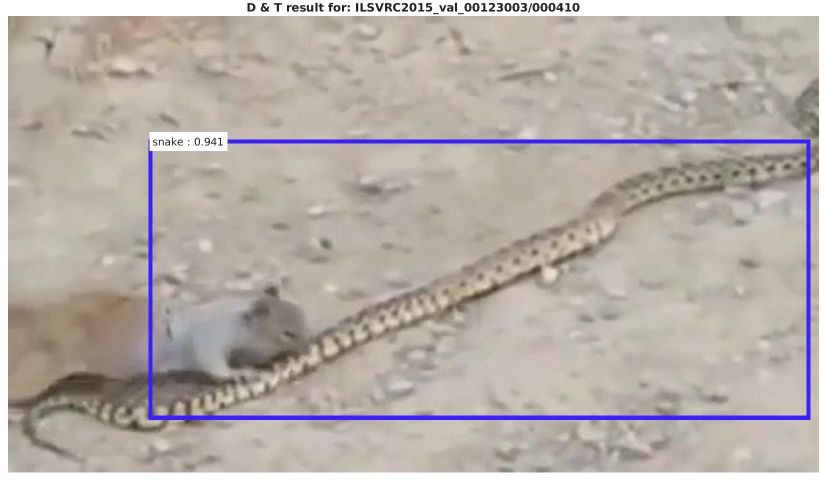} 
		\includegraphics[width=0.11\textwidth]{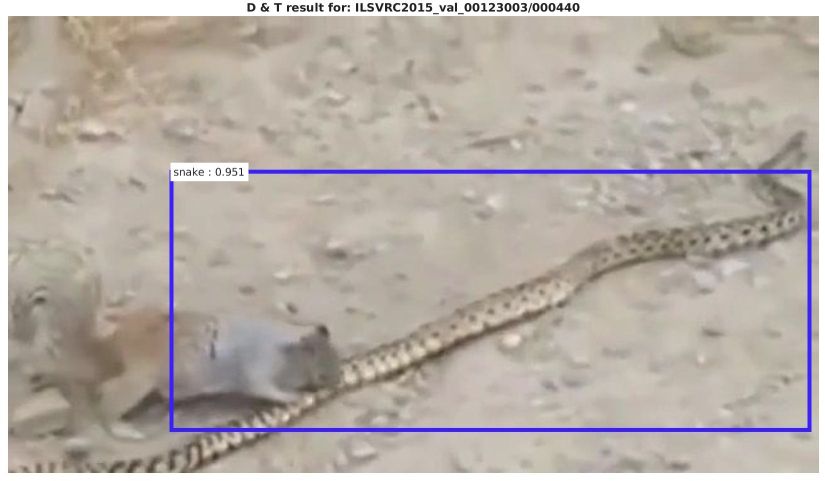}
		\includegraphics[width=0.11\textwidth]{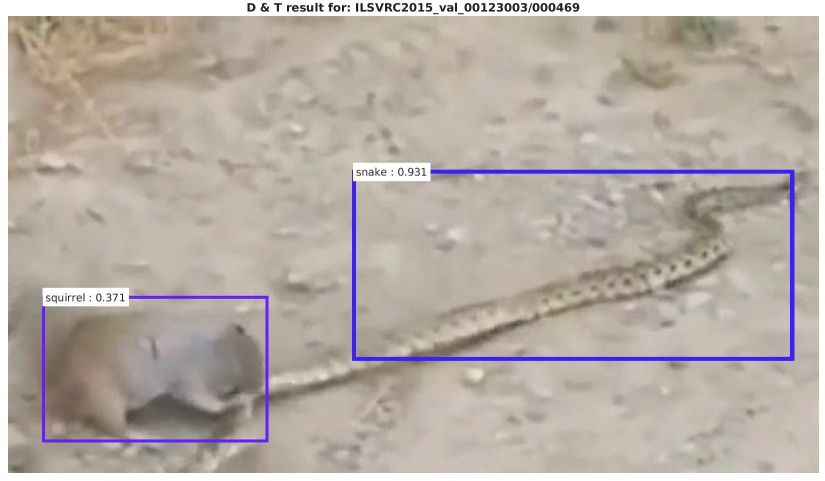}
		\includegraphics[width=0.11\textwidth]{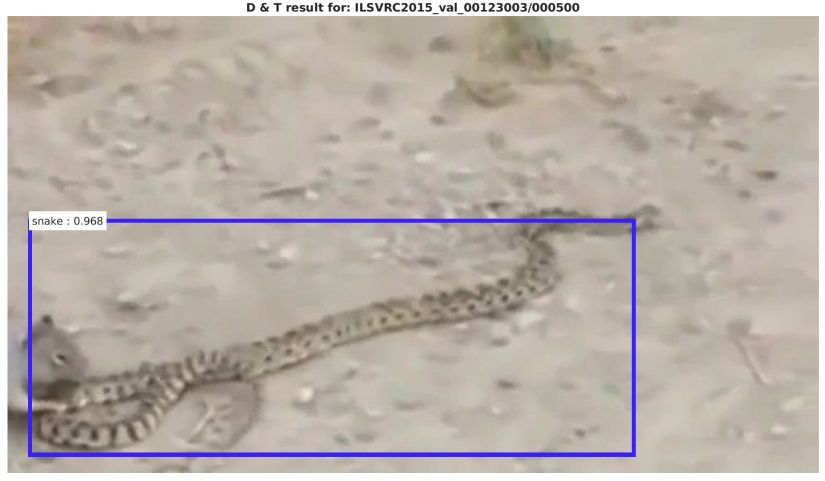}
	}

	\resizebox {1.0\textwidth }{!}{ 
		\includegraphics[width=0.11\textwidth]{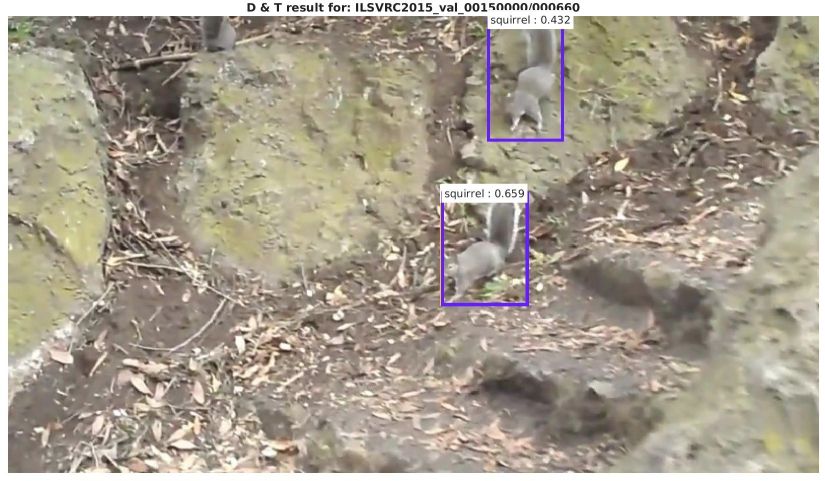}
		\includegraphics[width=0.11\textwidth]{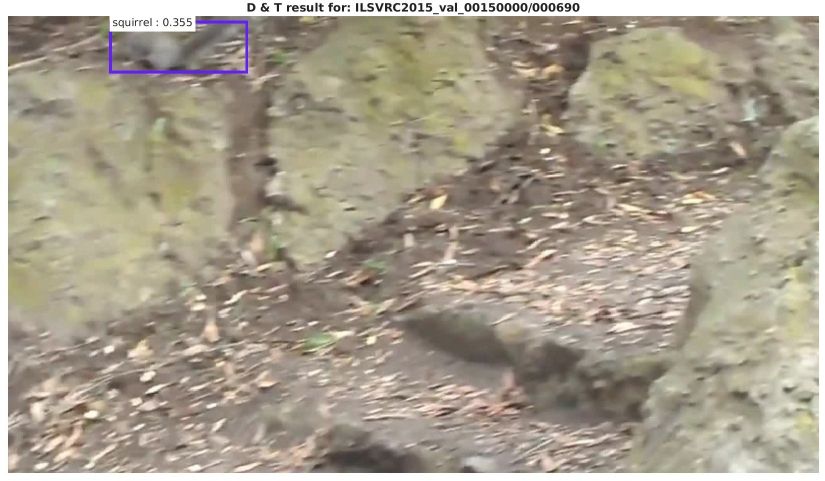}
		\includegraphics[width=0.11\textwidth]{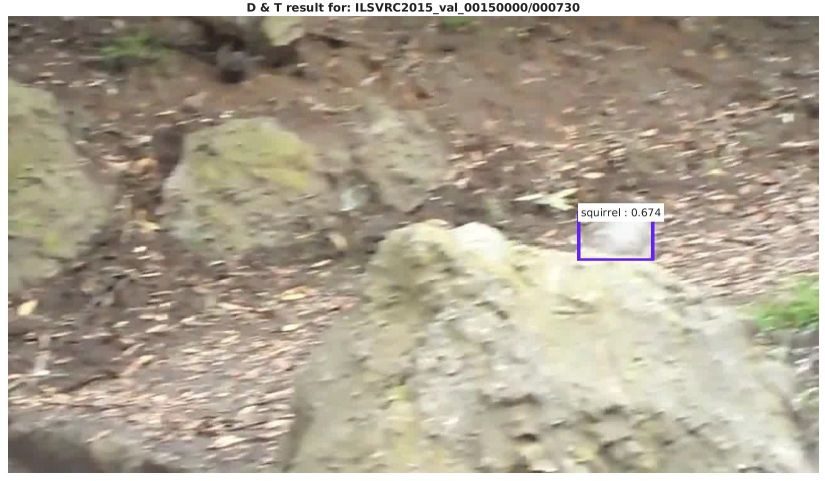}
		\includegraphics[width=0.11\textwidth]{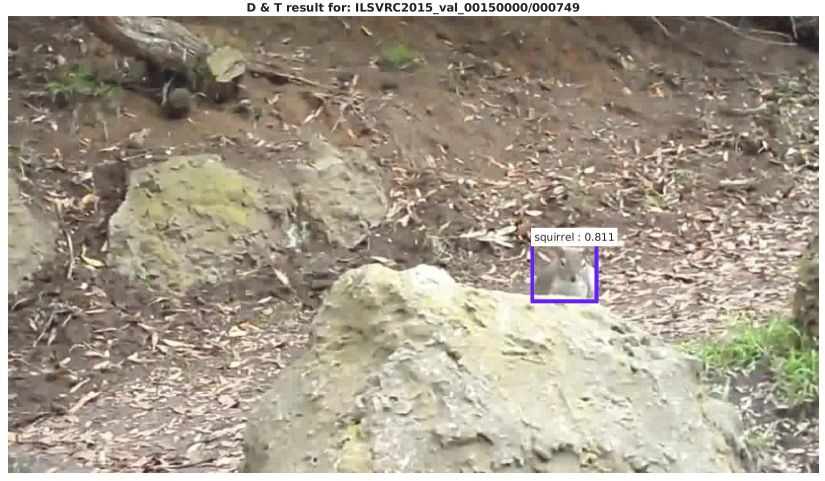}
	}
	\resizebox {1.0\textwidth }{!}{ 
		\includegraphics[width=0.11\textwidth]{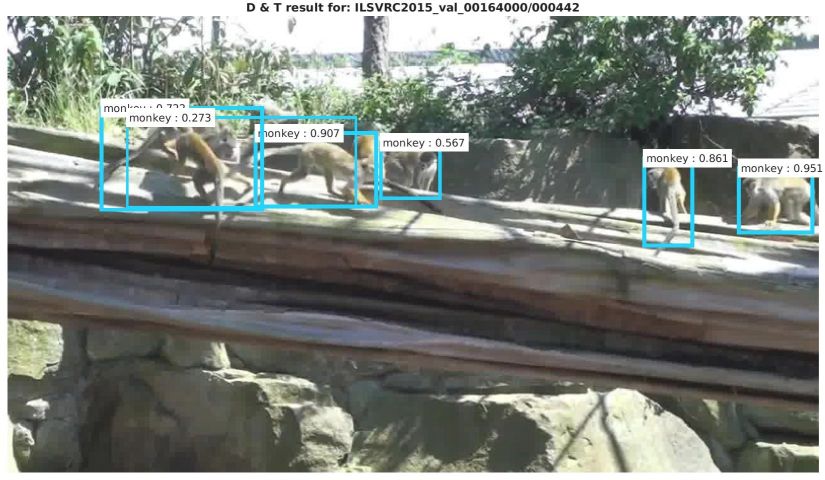}
		\includegraphics[width=0.11\textwidth]{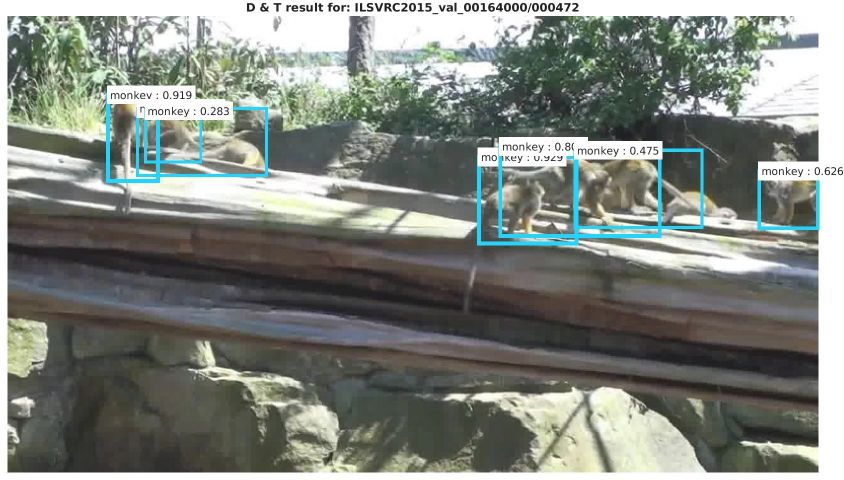}
		\includegraphics[width=0.11\textwidth]{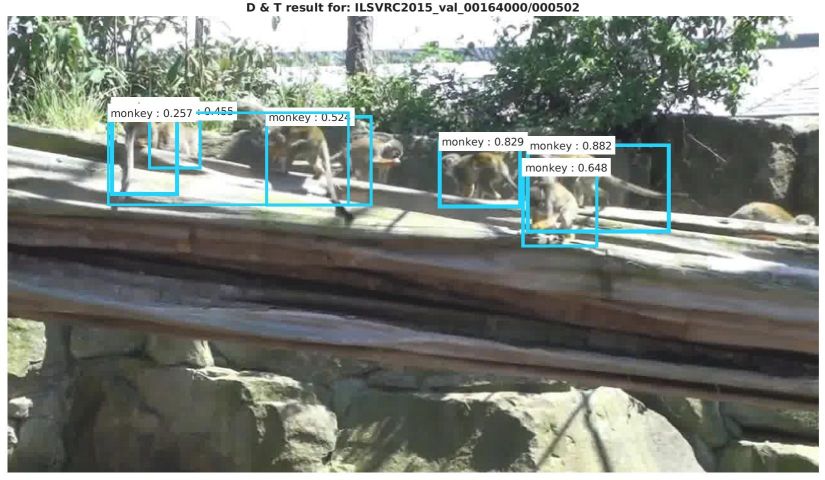}
		\includegraphics[width=0.11\textwidth]{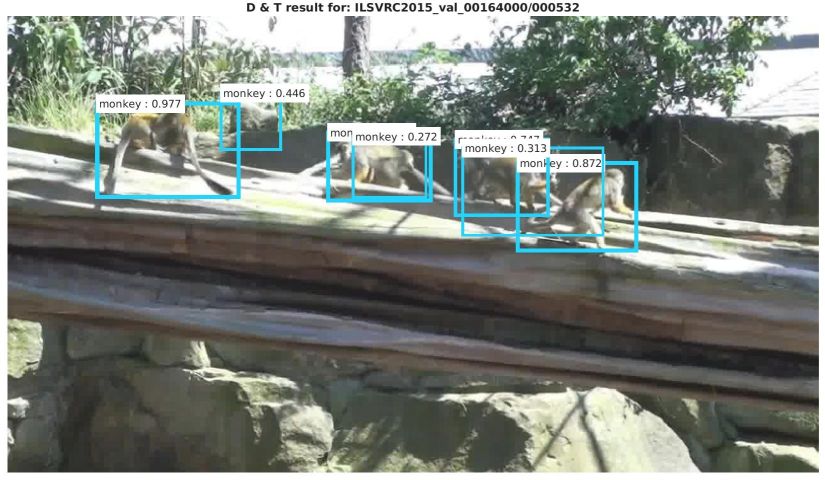}
	}
	\resizebox {1.0\textwidth }{!}{ 
		\includegraphics[width=0.11\textwidth]{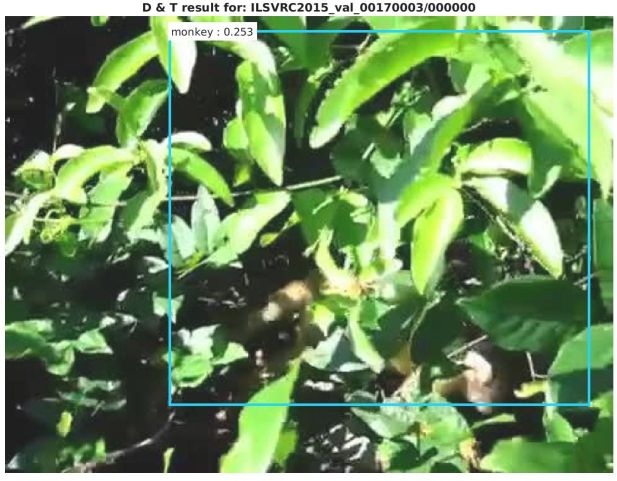}
		\includegraphics[width=0.11\textwidth]{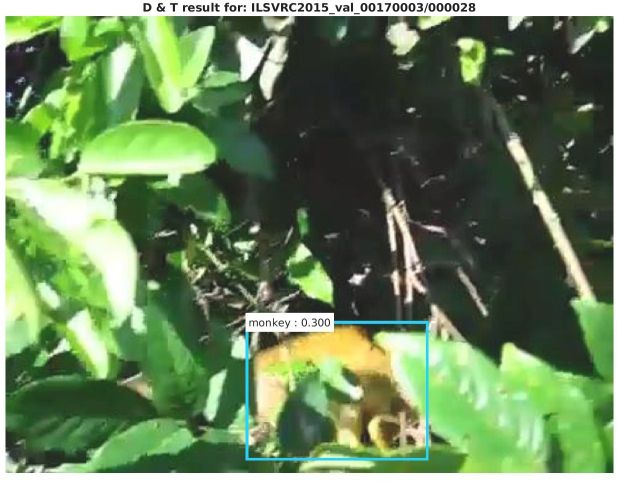}
		\includegraphics[width=0.11\textwidth]{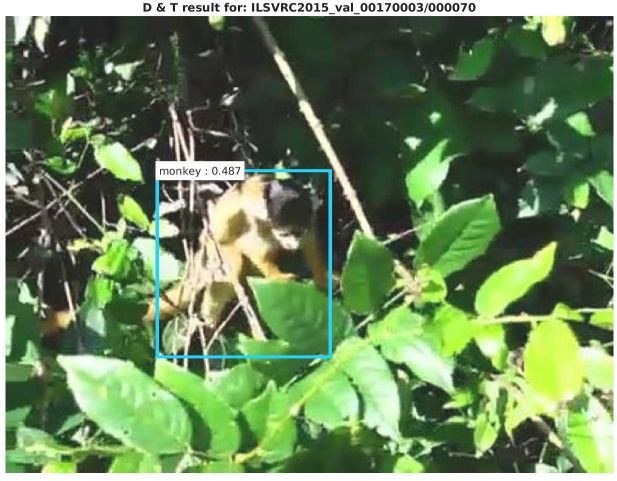}
		\includegraphics[width=0.11\textwidth]{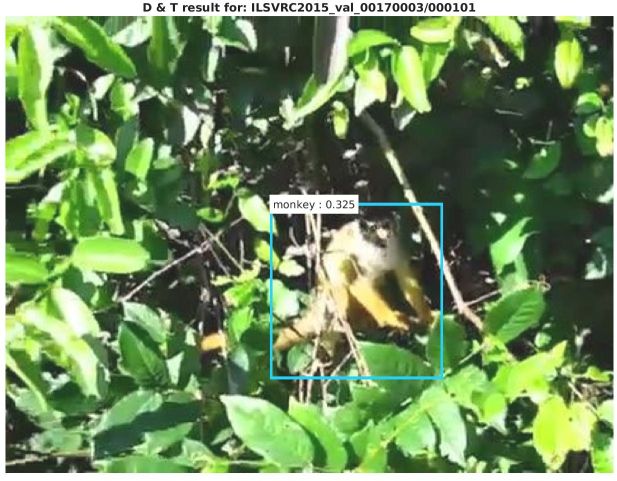}
	}
	\caption{Qualitative results for consecutive frames of videos where our D\&T approach could improve. Failures can be attributed to scale, occlusion, misclassification, or NMS issues.  }
	
	\vspace{-15pt}
	\label{fig:qualitative_results}
\end{figure*}

We show experimental results for our models and the current state-of-the-art in \tblref{tab:vid_val}. 
Qualitative results for difficult validation videos can be seen in \figref{fig:qualitative_results} and also at \url{http://www.robots.ox.ac.uk/~vgg/research/detect-track/}

\noindent\textbf{Frame level methods.}
First we compare methods working on single frames without any temporal processing. Our R-FCN baseline achieves 74.2\% mAP which compares favourably to the best performance of 73.9\% mAP in \cite{DFFcvpr17}. We think our slightly better accuracy comes from the use of 15 anchors for RPN instead of the 9 anchors in \cite{DFFcvpr17}. The Faster R-CNN models working as single frame baselines in \cite{kang2016object}, \cite{kang2017object} and \cite{kang2016tcnn} score with 45.3\%, 63.0\% and 63.9\%, respectively. We think their lower performance is mostly due to the difference in training procedure and data sampling, and not originating from a weaker base ConvNet, since our frame baseline with a weaker ResNet-50 produces 72.1\% mAP (\vs the 74.2\% for ResNet-101). Next, we are interested in how our model performs after fine-tuning with the tracking loss, operating via RoI tracking on the correlation and track regression features (termed D (\& T loss) in \tblref{tab:vid_val}). The resulting performance for single-frame testing is 75.8\% mAP. This 1.6\% gain in accuracy shows that merely adding the tracking loss can aid the per-frame detection. A possible reason is that the correlation features propagate gradients back into the base ConvNet and therefore make the features more sensitive to important objects in the training data. We see significant gains for classes like panda, monkey, rabbit or snake which are likely to move.  

\noindent\textbf{Video level methods.}
Next, we investigate the effect of multi-frame input during
testing. In \tblref{tab:vid_val} we see that linking our detections to
tubes based on our tracklets, D\&T ($\tau=1$), raises performance
substantially to 79.8\% mAP. Some class-AP scores can be boosted
significantly (\eg cattle by 9.6, dog by 5.5, cat by 6, fox by 7.9,
horse by 5.3, lion by 9.4, motorcycle by 6.4 rabbit by 8.9, red panda
by 6.3 and squirrel by 8.5 points AP). This gain is mostly for the
following reason: if an object is captured in an unconventional pose,
is distorted by motion blur, or appears at a small scale, the detector
might fail; however, if its tube is linked to other potentially highly
scoring detections of the same object, these failed detections can be
recovered (even though we use a very simple re-weighting of detections
across a tube). The only class that loses AP is whale ($-$2.6 points)
and this has an obvious explanation: in most validation snippets the whales
successively emerge and submerge from the water and our detection
rescoring based on tubes would assign false positives when they
submerge for a couple of frames.

When comparing our 79.8\% mAP against the current state of the art, we make the following observations. The method in \cite{kang2016object} achieves 47.5\% by using a temporal convolutional network on top of the still image detector. An extended work \cite{kang2017object} uses an encoder-decoder LSTM on top of a Faster R-CNN object detector which works on proposals from a tubelet proposal network, and produces 68.4\% mAP.
The ILSVRC 2015 winner \cite{kang2016tcnn} combines two Faster R-CNN detectors, multi-scale training/testing, context suppression, high confidence tracking \cite{wang2015visual} and optical-flow-guided propagation to achieve 73.8\%. And the winner from ILSVRC2016 \cite{ilsvrc16NUIST} uses a cascaded R-FCN detector, context inference, cascade regression and a correlation tracker \cite{ma2015hierarchical} to achieve 76.19\% mAP validation performance with a single model (multi-scale testing and model ensembles boost their accuracy to 81.1\%).

\noindent\textbf{Online capabilities and runtime.} The only component limiting online application is the tube rescoring (\sref{sec:linking}). We have evaluated an online version which performs only causal rescoring across the tracks. The performance for this method is 78.7\%mAP, compared to the noncausal method (79.8\%mAP). 
Since the correlation layer and track regressors are operating fully convolutional (no additional per-ROI computation is added except at the ROI-tracking layer), the extra runtime cost for testing a 1000x600 pixel image is 14ms (i.e. 141ms vs 127ms without correlation and ROI-tracking layers) on a Titan X GPU. The (unoptimized) tube linking (\sref{sec:linking}) takes on average 46ms per frame on a single CPU core).

\noindent\textbf{Temporally strided testing.}
We look at larger temporal strides $\tau$ during testing, which has recently been found useful for the related task of video action recognition \cite{feichtenhofer2016convolutional,feichtenhoferNIPS2016}. Our D \& T architecture is evaluated only at every $\tau^\text{th}$ frame of an input sequence and tracklets have to link detections over larger temporal strides. The performance for a temporal stride of $\tau=10$ is 78.6\% mAP which is 1.2\% below the full-frame evaluation. We think that such a minor drop is remarkable as the duration for processing a video is now roughly reduced by a factor of 10. 

A potential point of improvement is to extend the detector to operate over multiple frames of the sequence. We found that such an extension did not have a clear beneficial effect on accuracy for short temporal windows (\ie augmenting the detection scores at time $t$ with the detector output at the tracked proposals in the adjacent frame at time $t+1$ only raises the accuracy from 79.8 to \textbf{80.0\%} mAP). Increasing this window to frames at $t \pm 1$ by bidirectional detection and tracking from the $t^\text{th}$ frame did not lead to any gain. Interestingly, when testing with a temporal stride of $\tau=10$ and augmenting the detections from the current frame at time $t$ with the detector output at the tracked proposals at $t+10$ raises the accuracy from 78.6 to 79.2\% mAP. 

We conjecture that the insensitivity of the accuracy for short temporal windows originates from the high redundancy of the detection scores from the centre frames with the scores at tracked locations. The accuracy gain for larger temporal strides, however, suggests that more complementary information is integrated from the tracked objects; thus, a potentially promising direction for improvement is to detect and track over multiple temporally strided inputs.
\begin{table}[t]
	\centering
	\small{

		\begin{tabular}{l|c|c|c}
			\hline
			Backbone                        &  D  & D\&T   &  D\&T, average     \\	\hline
			
			ResNet-50      &          72.1	 &        76.5  &        76.7      \\
			\hline
			ResNet-101     &          74.1	&        79.8 &        80.0     \\
			\hline
			ResNeXt-101-32$\times$4  &  75.9 &        81.4	 &      81.6 \\
			\hline
			Inception-v4  &          77.9 &        82.0 &        \textbf{82.1} \\
			\hline
		\end{tabular}
	}
	\caption{Backbone network comparison for image-based \textbf{D}etection, and video-based \textbf{D}etection \& \textbf{T}racking  architectures. mAP (in \%) over all classes on ImageNet VID validation is shown.  }\vspace{-15pt}
	\label{tab:vid_val_backbones}
\end{table}

\noindent{\bf Varying the base network.}
Finally, we compare different base networks for the Detect \& Track architecture. \tblref{tab:vid_val_backbones} shows the performance for using 50 and 101 layer ResNets \cite{He16}, ResNeXt-101 \cite{xie2016aggregated}, and Inception-v4 \cite{szegedy2017inception} as backbones. We report performance for frame-level Detection (D), video-level Detection and Tracking (D\&T), as well as the variant that additionally classifies the tracked region and computes the detection confidence as the average of the scores in the current frame and the tracked region in the adjacent frame, (D\&T, average). We observe that D\&T benefits from deeper base ConvNets as well as specific design structures (ResNeXt and Inception-v4). The last row in \tblref{tab:vid_val} lists class-wise performance for D\&T with an Inception-v4 backbone that seems to greatly boost certain categories, \eg, dog (+5.7 AP), domestic cat (+9.4 AP) , lion (+11.4 AP), lizard (+4.5 AP), rabbit (+4.4 AP), in comparison to ResNet-101.

\section{Conclusion}
\label{sec:conclusion}

We have presented a unified framework for simultaneous object detection and tracking in video. Our fully convolutional D\&T architecture allows end-to-end training for detection and tracking in a joint formulation. In evaluation, our method achieves accuracy competitive with the winner of the last ImageNet challenge while being simple and efficient. We demonstrate clear mutual benefits of jointly performing the task of detection and tracking, a concept that can foster further research in video analysis.

\noindent\textbf{Acknowledgments.}
This work was partly supported by the Austrian Science Fund (FWF P27076) and by EPSRC Programme Grant Seebibyte 
EP/M013774/1.

{\small
\bibliographystyle{ieee}

\bibliography{DT_iccv17_arxiv}

}

\end{document}